\documentclass{article}

\usepackage[numbers, sort]{natbib}
\usepackage[final]{neurips_2025}

\usepackage[utf8]{inputenc} % allow utf-8 input
\usepackage[T1]{fontenc}    % use 8-bit T1 fonts
\usepackage{url}            % simple URL typesetting
\usepackage{booktabs}       % professional-quality tables
\usepackage{amsfonts}       % blackboard math symbols
\usepackage{nicefrac}       % compact symbols for 1/2, etc.
\usepackage{microtype}      % microtypography
\usepackage{xcolor}         % colors

\usepackage{pgf}
\usepackage{svg}
\usepackage{wrapfig}
\usepackage{caption}
\usepackage{amsmath, amsthm}
\newtheorem{lemma}{Lemma}

\newtheorem{proposition}{Proposition}

\usepackage{hyperref}       % hyperlinks
\usepackage[capitalize]{cleveref}
\usepackage{xspace}
\usepackage{xcolor}

\usepackage{algorithm}
\usepackage{algpseudocode}
\usepackage{amsmath, amssymb}

\usepackage[inline]{enumitem}

\newcommand{\quantile}[2]{\sQ\left(#1;#2\right)}

\newcommand{\parbold}[1]{\textbf{#1.}\xspace}
\usepackage{booktabs} 
\newcommand{\pertx}{\tilde{\vx}}
\newcommand{\Dplus}{\gD_{n+1}}
\newcommand{\Ecal}{\gE_{n}}
\newcommand{\Eplus}{\gE_{n+1}}

\usepackage{booktabs}
\usepackage{adjustbox}
\usepackage[table]{xcolor}
\usepackage{graphicx}
\usepackage{multirow}
\usepackage{subcaption}

\usepackage{amsmath,amsfonts,bm}

\def\eqref#1{equation~\ref{#1}}
\def\1{\bm{1}}

\def\vh{{\bm{h}}}

\def\vp{{\bm{p}}}
\def\vq{{\bm{q}}}

\def\vt{{\bm{t}}}

\def\vx{{\bm{x}}}

\def\vz{{\bm{z}}}
\def\vepsilon{\bm{\epsilon}}
\def\vdelta{\bm{\delta}}

\def\mI{{\bm{I}}}

\DeclareMathAlphabet{\mathsfit}{\encodingdefault}{\sfdefault}{m}{sl}
\SetMathAlphabet{\mathsfit}{bold}{\encodingdefault}{\sfdefault}{bx}{n}

\def\gA{{\mathcal{A}}}
\def\gB{{\mathcal{B}}}
\def\gC{{\mathcal{C}}}
\def\gD{{\mathcal{D}}}
\def\gE{{\mathcal{E}}}

\def\gH{{\mathcal{H}}}

\def\gL{{\mathcal{L}}}

\def\gN{{\mathcal{N}}}
\def\gO{{\mathcal{O}}}

\def\gR{{\mathcal{R}}}

\def\gU{{\mathcal{U}}}

\def\gX{{\mathcal{X}}}
\def\gY{{\mathcal{Y}}}

\def\sI{{\mathbb{I}}}

\def\sQ{{\mathbb{Q}}}
\def\sR{{\mathbb{R}}}

\newcommand{\E}{\mathbb{E}}

\newcommand{\method}{RCP1\xspace}
\newcommand{\dcal}{\gD_{n}}
\newcommand{\xtest}{\vx_{n+1}}
\newcommand{\xtestpert}{\tilde\vx_{n+1}}
\newcommand{\lowerval}{\mathrm{c}^\downarrow}
\newcommand{\upperval}{\mathrm{c}^\uparrow}

\newcommand{\differ}{\mathrm{d}}
\newcommand{\subautoref}[2]{\autoref{#1}-{#2}}

\title{One Sample is Enough to Make Conformal Prediction Robust}

\author{%
  Soroush H. Zargarbashi$^{1}$ \quad Mohammad Sadegh Akhondzadeh$^{2}$  \quad Aleksandar Bojchevski$^{2}$ \\
  $^{1}$ CISPA Helmholtz Center for Information Security, $^{2}$ University of Cologne \\
  \texttt{[zargarbashi, akhondzadeh, bojchevski]@cs.uni-koeln.de} \\
}

\begin{document}

\maketitle

\begin{abstract}
For any black-box model, conformal prediction (CP) returns prediction \emph{sets} guaranteed to include the true label with high adjustable probability. Robust CP (RCP) extends the guarantee to the worst case noise up to a pre-defined magnitude.
% enables it to tolerate worst-case perturbations up to a certain magnitude. 
%
For RCP, a well-established approach is to use randomized smoothing since it is applicable to any black-box model and provides smaller sets compared to deterministic methods. However, smoothing-based robustness requires many model forward passes per each input which is computationally expensive.
We show that conformal prediction attains some robustness even with \emph{a single forward pass on a randomly perturbed input}. Using any binary certificate we propose a single sample robust CP (\method). Our approach returns robust sets with smaller average set size compared to SOTA methods which use many (e.g. $\sim 100$) passes per input. 
Our key insight is to certify the conformal procedure itself rather than individual conformity scores. Our approach is agnostic to the task (classification and regression).  We further extend our approach to smoothing-based robust conformal risk control.
\end{abstract}

\section{Introduction}
\label{sec: Introduction}

Modern neural networks return uncalibrated probability estimates \citep{Guo2017OnCO}, and other uncertainty quantification methods (like Bayesian and ensemble models, Monte-Carlo dropout) are computationally expensive. Additionally, these methods do not usually provide formal statistical guarantees. Instead, conformal prediction (CP) is a post-processing method returning prediction \emph{sets} with a distribution-free and model-agnostic coverage guarantee, ensuring that the true answer is in the set with an adjustable high probability.
To apply CP, we need a conformity\footnote{Many works define CP via a non-conformity (disagreement) score. The setups are equivalent with a change in the score's sign. Our robustness results are invariant to this definition.} score function $s(\vx, y)$ capturing the agreement between $\vx$, and $y$ (e.g. softmax). We compute a conformal threshold over a holdout set of calibration points, and for the test points, we form the set as all labels with scores exceeding that threshold. These sets are guaranteed to include (cover) the true label with $1  - \alpha$ probability \citep{angelopoulos2024theoretical}. 
%

% COMMENT; I need to introduce the definition of score in this stage as I am using the upper/lower bound later in the introduction.
As shown in \subautoref{fig:first-fig}{left} (the red dashed line), this guarantee breaks by an unnoticeably small natural or adversarial noise to the test points -- the empirical coverage drastically decreases by an imperceptible perturbation. Note that from this point forward, we call adversarial or natural noises as ``perturbations'', not to be confused with the noise we (as the defender) introduce on purpose.
% Detailed descriptions for figures are in \autoref{sec:fig-manual}. \looseness=-1
% 
Robust CP (RCP) extends this guarantee to worst case bounded perturbations, ensuring that the perturbed input $\pertx$ is covered with the same or higher probability as the clean $\vx$, if $\vx$ is perturbed up to a known magnitude $r$ (e.g. $\|\pertx - \vx\|_2 \le r$). 
% Almost all RCP approaches are valid via this argument: if the clean point is covered (without accounting for perturbation) the perturbed point is also covered (with the conservative procedure). 
Previous RCP approaches find the highest/lowest possible conformity score within the perturbation ball, and replace the score with the worst-case bound\citep{gendler2021adversarially, jeary2024verifiably, zargarbashirobust, anonymous2024robust}. These bounds can be computed either analytically (Lipschitz bound or verifiers) or through randomized smoothing. Here, the trade-off is between the computational cost and the guaranteed robust radius -- analytical methods are robust to very smaller magnitudes of perturbation, while they only require a single forward pass per input. In contrast, while randomized smoothing needs many model forwards per single input, it provides robustness to significantly larger radii, and it applies over any black-box model.\looseness=-1

Smoothing is to augment the input with random noise and inference from the distribution of the model's output over smooth inputs instead.
% Randomized smoothing is extensively used for RCP since for nearby  the distribution of their smooth scores has a large overlap.
% Hence, these approaches redefine the score as a statistic of this distribution which changes slowly around $\vx$. 
For example, consider adding isotropic Gaussian noise $\vepsilon\sim\gN(\boldsymbol{0}, \sigma^2 \mI)$ to the input $\vx$, and defining the smooth score as the mean of the distribution $\E_{\vepsilon}[s(\vx + \vepsilon, y)]$. Regardless of the original model, the smooth model changes slowly near $\vx$, as the two distribution $\vx + \vepsilon$, and $\tilde\vx + \vepsilon$ have a large overlap. This leads to model-agnostic upper bounds on the score, around $\vx$. The upper bound is then used to decide whether a label is added to the prediction set.
% compared to the conformal threshold, deciding for the label to be in the set.
%model-agnostic lower bounds for this score, which can be used to change the acceptance threshold in CP to account for perturbations.
% Given any black-box score function, we define a smooth score which looks at the distribution of the scores at random points around the input. Regardless of the score function itself, smooth scores change slowly around the input which enables us to provide tighter lower bounds and therefore the resulting conformal procedure needs to be less conservative. 
An important drawback here is the computational overhead. The score function (e.g. the expectation of the smooth scores) must be estimated via Monte-Carlo sampling. By reducing the number of samples, the confidence intervals for the mean widen, and the size of the prediction sets quickly increases. 

% An efficient alternative is to use neural network verifiers \citep{jeary2024verifiably}, however, they are limited to simpler models, and small radii. The question is:  \looseness=-1 
We answer this question: \emph{Can we design smoothing-based RCP without introducing computational overhead?}
Interestingly, we show the vanilla CP combined with noise-augmented inference already has robust behavior (green dashed line in \subautoref{fig:first-fig}{left}). 
% This allows us to directly apply robustness certificates on CP's guarantee -- the fixed value $1 - \alpha$, which in contrast to mean, CDF, or quantile-based scores used in other RCPs, does not need to be estimated via sampling.
With that, we define \method (\underline{r}obust \underline{c}onformal \underline{p}rediction with \underline{one} sample) that returns robust sets using a single inference per point. In practice, the resulting sets have the same guarantee, and similar size, to the previous SOTA which needs around 100 samples per input instead.
% that require many forward passes per input. For instance, \method returns sets similar to BinCP \citep{zargarbashirobust} which needs at least 70 to 110 samples per input.
By nullifying the need for sampling, we can use even larger models (like vision transformers) to return even smaller sets (see \subautoref{fig:first-fig}{middle}). Importantly, we do not compete with the SOTA equipped with unlimited sampling (compute) budget. Instead, we propose a compute-friendly alternative that still produces small prediction sets in regimes infeasible for the other smoothing-based RCP methods (large models and limited computational power).\looseness=-1

\begin{figure}
    \centering
    \input{figures/first-page-figure.pgf}
    \caption{[Left] Coverage of vanilla CP, our robust RCP1,
    and the SOTA BinCP under adversarial attack (\autoref{sec:fig-manual}). $\gC_r$ denotes sets with robustness guarantee up to radius $r$, and $C_0$ is guaranteed only for clean points -- but still using same process only with $r = 0$. Even sets from $\gC_0$ show significantly higher coverage compared to vanilla due to randomization.
    [Middle]
    % the difference in the average set size of BinCP (SOTA) and \method expressed as  $|\gC_{r, \mathrm{BinCP}}| - |\gC_{r, \mathrm{\method}}|$. The plot compares over various sample rates (only used by BinCP) and radii. Both plots are for CIFAR-10 dataset and \texttt{ResNet}. [Right] 
    The average time to compute, and the average set size for both \method and BinCP, with ResNet and ViT models on the ImageNet dataset; both axis are log-scaled and pareto-optimal points are at the lower-left. RCP1 is more efficient. Both plots are with $\sigma=0.5$.
    [Right] Smoothing-based robust conformal risk control. We show the coverage and miscoverage of the RCP1 mask for the class "car" in the segmentation task. Here risk is set to false negative rate.}
    \label{fig:first-fig}
\end{figure}

% \method is easy to implement (see \autoref{alg:split-conformal}), agnostic to the model, the distribution of inputs, and score function: use the smoothing scheme (e.g. isotropic Gaussian noise) to augment each datapoint, and run the conformal calibration similar to the vanilla setup but instead with a conservative $1 - \alpha'$ nominal coverage (from \autoref{sec: Robust CP with a Single Sample}). We choose $1 - \alpha'$ such that our certified lower bound remains above $1 - \alpha$ using any binary certificate (see \autoref{sec: Extension to All Shapes and Sizes} for various smoothing and threat models). Interestingly, our guarantee is task independent (classification, or regression).
\method is similar to vanilla CP only with two changes: \begin{enumerate*}[label=(\roman*)]
    \item we use noise-augmented input (from the smoothing) to compute the scores, and \item we calibrate with a conservative $1 - \alpha'$ nominal coverage chosen such that our certified lower bound (in \autoref{sec: Robust CP with a Single Sample}) remains above $1 - \alpha$. 
\end{enumerate*} \method works with any binary certificate (see \autoref{sec: Extension to All Shapes and Sizes}), is agnostic to the model, the distribution of inputs, and score function, and  interestingly, it is task independent -- same binary certificate works for both classification, or regression.
% In \autoref{sec: Extension to All Shapes and Sizes}, we extend \citet{yang2020randomized}'s classification certificates for robust CP to apply for various smoothing schemes and threat models.
We use a similar process to define a smoothing-based conformal risk control (\subautoref{fig:first-fig}{right}).

\section{Background}
\label{sec: Background}

CP requires a holdout set of labeled calibration points $\dcal = \{\vx_i, y_i\}_{i = 1}^n$ that are exchangeable with the future test point $\vx_{n+1}$. From the model's output, we define a score function $s:\gX \times \gY \rightarrow \sR$ where it quantifies the agreement between $\vx$, and $y$, e.g. softmax; see \autoref{sec: More on Conformal Prediction} for other scores. \citet{Vovk2005AlgorithmicLI} show that under exchangeability (vanilla setup) the set $\gC_0(\vx_{n+1}) = \{y: s(\vx_{n+1}, y) \ge q\}$ for $q = \quantile{\alpha}{\{s(\vx_i, y_i): (\vx_i, y_i) \in \dcal\}}$ contains the true label $y_{n+1}$ at least with $1 - \alpha$ probability.\begin{align}
\label{eq:conformal-guarantee}
\Pr_{\Dplus}[y_{n+1} \in \gC_0(\xtest)] = \Pr_{\Dplus}[s(\vx_{n+1}, y_{n+1}) \ge q] \ge 1 - \alpha
\end{align}
Here $\Dplus = \dcal \cup \{(\vx_{n+1}, y_{n+1})\}$, and $\quantile{\alpha}{\gA}$ is the $\lfloor\alpha\cdot(1 - \frac{1}{n})\rfloor$ quantile of the set $\gA$.
While the coverage guarantee is agnostic to the model (and the score), better model or score functions reflects in properties like the prediction set size ( a.k.a efficiency). 
While methods like \citep{zargarbashirobust} require bounded score, our results are also agnostic to the choice of the score function (bounded or unbounded).
% We discuss other score function later in \autoref{appendix}.

\parbold{Threat model}
We consider the worst case (or adversarial) perturbation, which yields a more powerful guarantee compared to probabilistic robustness e.g. from \citet{Ghosh2023ProbabilisticallyRC}. 
% With (worst-case) perturbations, the exchangeability
% between the calibration and test points
% breaks and the empirical coverage drastically decreases from the guaranteed level (see \subautoref{fig:first-fig}{left}).
% 
In our threat model, the adversary aims to decrease the empirical coverage below the guaranteed $1 - \alpha$ by adding an imperceptible noise to the test points (evasion). The set of all possible perturbations is defined as a ball $\gB:\gX\rightarrow 2^\gX$ around the clean input. We define an inverted ball $\gB^{-1}$ as the smallest set that contains the original (clean) point from any possible perturbation; i.e. $\forall \tilde\vx \in \gB(\vx) \Rightarrow \vx \in \gB^{-1}(\tilde\vx)$. For images, a common threat model is $\ell_2$-norm: $\gB_r(\vx) = \{ \tilde\vx: \|\tilde\vx - \vx\|_2 \le r\}$ where $r$ is the radius of the perturbation. For symmetric balls like $\ell_2$ we have $\gB = \gB^{-1}$, but this does not hold in general (e.g. \citep{bojchevski2020efficient}). \looseness=-1

\parbold{Robust Conformal Prediction (RCP)} Robust CP extends the guarantee in \autoref{eq:conformal-guarantee} to the worst case noise. For $\gB$, prior works define a robust (conservative) prediction set $\gC_\gB$ satisfying the following \begin{align}
    \label{eq:robust-conformal-guarantee}
    \Pr_{\Dplus}[y_{n+1} \in \gC_\gB(\xtestpert), \forall \xtestpert \in \gB(\xtest)] \ge 1 - \alpha
\end{align}
% Leaving the details to \autoref{sec: Robust CP with a Single Sample} and \autoref{sec: Threat Model}, in brief, 
\autoref{eq:robust-conformal-guarantee} is only meaningful 
% for RCPs that produce
for deterministic sets. We discuss this subtle point in \autoref{sec: Robust CP with a Single Sample} and \autoref{sec: Threat Model}. Earlier smoothing-based RCP methods implicitly remove all inherent randomness, which makes the definition applicable to them. These methods can be summarized with the following two arguments:
\begin{enumerate*}[label=(\roman*)]
    \item for the exchangeable $\xtest$ CP covers the true label with $1 - \alpha$ probability, \item if the clean $\xtest$ was originally covered by (vanilla) CP, robust CP also covers $\xtestpert$, because if a clean score is above $q$ its upper bound (over any perturbed input) is also above $q$ \citep{anonymous2024robust}. Thus, the vanilla set for $\xtest$ is a subset of the robust set for $\xtestpert$.   
\end{enumerate*} This results in robust coverage of \emph{at least} $1 - \alpha$. 
To account for the inherent randomness in CP, in \autoref{sec: Robust CP with a Single Sample}, we redefine the threat model, and replace the argument (ii) by the following: ``the perturbed $\xtestpert$ has a higher probability to be in the robust prediction set compared to $\xtest$ being in the vanilla set''. The new formulation still addresses the worst-case perturbation.

\parbold{Certified bounds} For any function $f$ and ball  $\gB(\vx)$ define the certified lower bound as $\lowerval[f, \vx, \gB] \le \inf\{ f(\vz) : \vz \in \gB(\vx) \}$, and similarly $\upperval[\cdot, \cdot, \cdot]$ as the upper bound (with $\sup$). With this definition, for each $\vx$ we have $\forall \tilde\vx \in \gB(\vx), \lowerval[s(\cdot, y), \vx; \gB] \le s(\tilde\vx, y) \le \upperval[s(\cdot, y), \vx, \gB]$ where we plug in the score function for $f$. \citet{zargarbashirobust} show that given these certified bounds within $\gB$, the conservative sets defined either as  $\gC_\gB(\vx_{n+1}) = \{ y: \upperval[s(\cdot, y_{n+1}), \vx_{n+1}; \gB^{-1}] \ge q\}$ (test-time RCP), or similarly, $\gC_\gB(\vx_{n+1}) = \{ y: s(\xtest, y) \ge \bar{q}\}$ for $\bar{q} = \quantile{\alpha}{\{\lowerval[s(\cdot, y_{i}), \vx_{i}; \gB]: (\vx_i, y_i) \in \dcal\}}$ (calibration-time RCP) attain $1 - \alpha$ robust coverage. 
% Here $\lowerval[f, x, \gB]$ is the certified lower bound for the function $f$ over any point within $\gB(\vx)$; $\upperval$ is similarly the upper bound.
%
% Note that for the test-time RCP we bound the score w.r.t. $\gB^{-1}$ -- for the potentially perturbed input we bound the score of the clean point. Oppositely, in the calibration-time RCP we find the worst perturbation given the clean input. \looseness=-1

\parbold{Randomized smoothing}
One approach to compute these upper/lower bounds for any black-box model, or score is randomized smoothing. A smoothing scheme $\xi: \gX \rightarrow \gX$ adds a random noise to the input --  maps it to a random point close to it. 
A common smoothing for continuous data (e.g. images) is the Gaussian smoothing $\xi(\vx) = \vx + \vepsilon$ where $\vepsilon$ is an isotropic Gaussian noise $\vepsilon \sim \gN(\boldsymbol{0}, \sigma^2\mI)$. While our method works for any smoothing, for easier notation we further use $\vx + \vepsilon$ instead of $\xi(\vx)$. \looseness=-1 

For any score function $s$, the distribution of the smooth scores $s(\vx + \vepsilon, y)$ changes slowly. This enables us to compute tight bounds on the smooth statistics (mean, quantile, etc.) within $\gB$, or $\gB^{-1}$. RSCP \citep{gendler2021adversarially, yan2024provably}, and CAS \citep{zargarbashirobust} set the score function directly to the mean of the distribution. BinCP \citep{anonymous2024robust}, uses the $p$-quantile instead. 
These statistics are often intractable to compute and therefore estimated using Monte-Carlo sampling, followed by a finite sample correction. \method however nullifies the need to estimate these statistics. We discuss the related work further in \autoref{sec:related-works}.

\begin{figure}[t]
    \centering
    \begin{minipage}{0.5\textwidth}
        \includegraphics{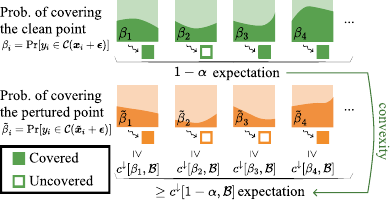}
        \caption{Illustration of the theory behind \method. The probabilities $\beta_i$ never need to be computed, since due to the convexity of $\lowerval$ we can directly work with $1 - \alpha$. Further description in \autoref{sec: Robust CP with a Single Sample}.}
        \label{fig:illustration}
    \end{minipage}
    \begin{minipage}{0.48\textwidth}
  \captionsetup{type=algorithm}       % make caption/numbering be "Algorithm"
  \refstepcounter{algorithm}          % increment the Algorithm counter for labels/refs

  % --- ruled header like the algorithm float ---
  \noindent\rule{\linewidth}{0.6pt}\vspace{0.5pt}
  \textbf{Algorithm~\thealgorithm.} \,%
  \method; the \textcolor{teal}{colored} part shows the difference with vanilla CP.
  \label{alg:split-conformal}
  \vspace{-2pt}\par
  \noindent\rule{\linewidth}{0.4pt}\vspace{1pt}

  % --- the algorithmic body (optionally smaller font) ---
  \begingroup\small
  \begin{algorithmic}[1]
    \Require Calibration set $\dcal=\{(\vx_i,y_i)\}_{i=1}^n$; nominal coverage $1-\alpha\in(0,1)$; score $s:\mathcal{X}\times\mathcal{Y}\to\mathbb{R}$; potentially perturbed test point $\xtestpert$
    \State Compute $s_i \gets s\!\left(\vx_i \textcolor{teal}{+ \vepsilon},\,y_i\right) : (\vx_i, y_i)\in\mathcal{D}$.
    \State \textcolor{teal}{Set} $\textcolor{teal}{1 - \alpha' \gets \upperval[1 - \alpha, \mathcal{B}^{-1}]}$ \Comment{e.g., Gaussian smoothing with $\gB_r$: $\Phi_\sigma(\Phi_\sigma^{-1}(1-\alpha)+r)$.
    % See \autoref{sec:choosing-conservative}.
    }
    \State Set $\bar{q}_\alpha = \sQ\left(\textcolor{teal}{\alpha'},\{s_i\}_{i=1}^n\right)$.
    \State For input $\xtestpert$ \Return \[\mathcal{C}_r(\xtestpert)=\{y: s(\xtestpert\textcolor{teal}{+\vepsilon},y)\ge {\bar{q}_\alpha}\}\]
  \end{algorithmic}
  \endgroup

  \vspace{0pt}
  \noindent\rule{\linewidth}{0.6pt}   % bottom rule
\end{minipage}
\end{figure}

\section{\method: Robust CP with One Sample}
\label{sec: Robust CP with a Single Sample}

\parbold{High level view} We prove that when the scores are computed on noise-augmented inputs, i.e. using $s(\vx + \vepsilon, y)$ instead of $s(\vx, y)$ for both calibration and prediction (see \autoref{alg:split-conformal}), vanilla CP already yields robust prediction sets - and its coverage under perturbation can be bounded. We provide a sketch of our arguments. To prove it 
% \begin{enumerate*}[label=(\roman*)]
    % \item 
    we first assume an abstract value $\beta_{n+1}$ as the coverage probability of a specific clean point $\xtest$, which is over the random noise and the inherent randomness in the score.
    % \item 
    With $\tilde{\beta}_{n+1}$ as the coverage probability for $\xtestpert$ (again over augmented input) we show that $\tilde\beta_{n+1}$ can be lower bounded using the randomized smoothing certificates.
    % Here we certify the abstract value $\beta_{n+1}$ which is assumed without estimation.
    % \item 
    By showing the convexity of the certificate w.r.t. $\beta_{n+1}$, we can directly lower bound the expected coverage over all inputs which is $1 - \alpha$. We never need to compute the abstract $\beta_{n+1}$ or $\tilde\beta_{n+1}$.
% \end{enumerate*}
This sketch is illustrated in \autoref{fig:illustration}.

\parbold{Limitation of \autoref{eq:robust-conformal-guarantee}}  
The universal quantifier in \autoref{eq:robust-conformal-guarantee} implies that for $\ge 1-\alpha$ fraction of test points, \emph{all} $\xtestpert$ must be deterministically covered, including the clean $\xtest$.
However, many CP methods (like APS \citep{angelopoulos2024theoretical}) incorporate internal stochasticity (e.g. to break ties), making the coverage event a random variable rather than a binary indicator. This is true even before we add our noise on top. Hence, \autoref{eq:robust-conformal-guarantee} does not reduce to \autoref{eq:conformal-guarantee} for $r = 0$. For random sets, adversary can reduce the coverage \emph{probability} for each $\xtest$. With $u$ encoding all the (inherent) randomness in the sets, $\gC_0$ as the vanilla set, and $\gC_r$ as the robust set for a ball $\gB_r$, we rewrite the guarantee as:
% \[\min_{\xtestpert \in \gB_r(\xtest)} \Pr[y_{n+1} \in \gC_r(\xtestpert)] \ge \Pr[y_{n+1} \in \gC_0(\xtest)] \ge 1 - \alpha\]
\begin{align}
    \label{eq:correct-guarantee}
    \Pr_{\Dplus}[\min_{\xtestpert \in \gB_r(\xtest)} \Pr_{u}[y_{n+1} \in \gC_r(\xtestpert; u)]] \ge \Pr_{\Dplus, u}[y_{n+1} \in \gC_0(\xtest; u)] \ge 1 - \alpha
\end{align} 

% Since \autoref{eq:robust-conformal-guarantee} does not account for the inherent randomness of the CP itself, it is vague for random scores (like APS - see \autoref{sec: More on Conformal Prediction}) as in that setup $y_{n+1} \in \gC(\xtest)$ is not binary. With the randomness in coverage of each point, the adversary's goal is to decrease the coverage probability over internal randomness of the process. 

% Here, $\gC_0$ is the vanilla and $\gC_r$ is the robust set for a ball $\gB_r$ 

% We use a random score function which we can quantify what is the worst case 

% In the previous robust CP methods, for a realisation of $\dcal$, and $\xtest$, the first argument was deterministic; e.g. in CAS \citep{zargarbashirobust} $\sI[\hat{s}(\xtest, y_{n+1}) \ge q_0]$ is either 1 or 0 for known test and calibration points. Despite previous methods our conformal sets are probabilistic even in the vanilla setup. 

\parbold{General worst-case guarantee} We prove the lower bound coverage guarantee in \method, as implemented in \autoref{alg:split-conformal}. We state the result in terms of random variables, which have a specific realization in practice. Let $Z_i: i \in [n+1]$ be $n+1$ exchangeable random variables, where $Z_i = (X_i, Y_i)$ for $X_i \in \gX$ (e.g. $\gX = \sR^d$), and $Y_i \in \gY$ (e.g. $\gY = [K]$). 
% We can think of $(\vx_i, y_i)$'s as realizations of these random variables. 
Let $s:\gX\times\gY\to\sR$ be any measurable score function. Let $E_i:i\in[n+1]$ be i.i.d. random variables from a distribution supported on $\gX$ (e.g. $E_i \sim \gN(\boldsymbol{0}, \sigma^2\mI)$). We define $\hat{S}_i = s(X_i, Y_i)$, and $S_i = s(X_i + E_i, Y_i)$. 
% Note that due to frequency \emph{we used} $\hat{S}_i$ \emph{as the vanilla scores}.
Let $\delta \in \gB_r$ be any arbitrary perturbation up to radius $r$, define $\tilde{X}_{n+1} = X_{n+1} + \delta$, and $\tilde{S}_{n+1} = s(\tilde{X}_{n+1} + E_{n+1}, Y_{n+1})$ accordingly.\looseness=-1 
% In the rest, we prove the following:
\begin{proposition}
    \label{thrm:rcp1}
    Let $q = \quantile{\alpha}{s(X_i + E_{i}, Y_i): (X_i, Y_i) \in \dcal\}}$. Given a certified lower bound $\lowerval[\cdot, \gB]$ as later defined in \autoref{eq:certificate-pointwise}, and $\Eplus=\{E_i\}_{i=1}^{n+1}$, for any perturbation $\delta \in \gB_r$ we have \[
    \Pr_{\Dplus, \Eplus}[s(\tilde{X}_{n+1} + E_{n+1}, Y_{n+1}) \ge q] \ge \lowerval[1- \alpha, \gB]
    \] 
\end{proposition}
\begin{proof}
Adding i.i.d. noise $E_i$ is permutation equivariant, thus using $S_i$'s 
% instead of $\hat{S}_i$
doesn't break exchangeability \cite{angelopoulos2024theoretical}.
From \citet{Vovk2005AlgorithmicLI} 
we have $
    \Pr[S_{n+1} \ge q] \ge 1 - \alpha$ for $q = \quantile{\alpha}{\{S_i\}_{i = 1}^n}$,
where the probability is over $\Dplus$, and $\Eplus$.
% \footnote{Moreover, for a fixed calibration set $\dcal = \{Z_i\}_{i = 1}^n$, 
% % and every draw of $E_i$'s
% under additional mild assumptions:
% $
% \Pr[S_{n+1} \ge q \mid \dcal] \sim \mathrm{Beta}((1-\alpha)\cdot(n+1), \alpha(n+1))
% $.
% } 
We can rewrite this as 
% This is formally written as
% Importantly if we marginalize out all the $E_{i}$'s we still have
% $\Pr_{\Dplus, \Eplus}[S_{n+1} \ge q] \ge 1-\alpha$.
% We can rewrite
\[
\Pr_{\Dplus, \Eplus}[S_{n+1} \ge q] 
=
\E_{\Dplus,  \Ecal} \left[ \Pr_{E_{n+1}}[S_{n+1} \ge q ] \mid \Dplus,  \Ecal \right]
=
\E_{\Dplus,  \Ecal}  [ \beta_{n+1} ]
\ge 1-\alpha
\]
where we define $\beta_{n+1}:= \Pr_{E_{n+1}}[S_{n+1} \ge q \mid \Dplus, \gE_n]$. Here, $\beta_{n+1}$ is only a probability over the last noise $E_{n+1}$ (and any other internal randomness in the score) for a fixed $\Dplus$ and $\gE_n$. We call $\beta_{n+1}$ the clean instance-wise coverage. Similarly for $\tilde{X}_{n+1}$ we define $\tilde\beta_{n+1}$.

%
% By marginalizing over the space of all possible $\dcal$'s, and $E_i$'s we get the marginal coverage guarantee $\E_{\Dplus}[\E_{E_i:i \in [n]}[\beta \mid \Dplus, E_i:i\in[n]]] \ge 1 - \alpha$. 
%
We can bound any smooth binary function with an existing certified lower bound $\lowerval[\cdot, \gB]$. Formally,
\begin{align*}
    \Pr_{E_{n+1}}[S_{n+1} \ge q \mid \Dplus, \gE_n] = \beta_{n+1} \quad \Rightarrow \quad \Pr_{E_{n+1}}[\tilde{S}_{n+1} \ge q \mid \Dplus, \gE_n] \ge \lowerval[\beta_{n+1}, \gB]
\end{align*} 
Note that here both $\beta_{n+1}$ and $\tilde\beta_{n+1}$ share the same $\Dplus$, and $\gE_n$ and both are over the random variable $E_{n+1}$ and the inherent randomness of the score. Later in \autoref{thrm:certificate-convex}, we show that the function $\lowerval[\beta, \gB]$ is convex and increasing in $\beta$. This helps us to bound the adversarial coverage guarantee as \begin{align*}
    \Pr[Y_{n+1} \in \gC(\tilde{X}_{n+1})] = \Pr[\tilde{S}_{n+1} \ge q] &= \E_{\Dplus, \gE_n}[
     \Pr_{E_{n+1}}[\tilde\beta_{n+1}]] \\ 
    % \Pr_{E_{n+1}}[\tilde{S}_{n+1} \mid \Dplus, \gE_n]] \\ 
    \textcolor{gray}{\text{(from certificate)}}
    &\ge \E_{\Dplus, \gE_n}[\lowerval[\beta_{n+1}, \gB]]
    % = \E_{\Dplus}[\E_{\gE_n}[\lowerval[\beta, \gB]]] 
    \\
    \textcolor{gray}{(\text{from  convexity (\autoref{thrm:certificate-convex})}, \E[\lowerval[\beta, \cdot]] \ge \lowerval[\E[\beta], \cdot]])} &\ge \lowerval[\E_{\Dplus, \gE_n}[\beta_{n+1}], \gB] \ge \lowerval[1 - \alpha, \gB]
\end{align*}
where the last inequity holds due to vanilla CP and monotonicity.
\end{proof}
% also we can switch the expectations due to Fubini's theorem.
% For the clean points the coverage guarantee holds both marginally and conditional to $\Eplus = \gE_n \cup E_{n+1}$.
Just like with any other CP with any kind of randomness in the score (e.g. APS), the guarantee only holds marginally over $\Eplus$ and the internal randomness. In other words, the coverage probability is higher than $\lowerval[\beta_{n+1}, \gB]$ for specific $\tilde X_{n+1}$, 
and $\lowerval[1 - \alpha, \gB]$ on average, if we draw a random $\Eplus$. If we instead fixed $\Eplus$ and the adversary knows the noise, the guarantee can easily break.
Note that $\beta_{n+1}$ is an abstract quantity, the probability that $X_{n+1} + E_{n+1}$ is covered. In principle we can not estimate $\beta_{n+1}$, since the label is not known. Nonetheless, due to the convexity, we can lower bound the coverage guarantee directly without that information.

\parbold{Instance-wise worst case coverage}
\autoref{thrm:rcp1} relies on lower bounding the worst-case (adversarial) $\tilde\beta_{n+1} = \Pr_{\vepsilon_{n+1}}[s(\xtestpert + \vepsilon_{n+1}, y_{n+1}) \ge q]$, for the perturbed $\xtestpert = \xtest + \vdelta \in \gB(\xtest)$ given $\beta_{n+1} := \Pr_{\vepsilon_{n+1}}[s(\xtest + \vepsilon_{n+1}, y_{n+1}) \ge q]$. Here, $(\vx_{n+1}, y_{n+1})$ and $\vepsilon_{n+1}$ are realizations of $(X_{n+1}, Y_{n+1})$ and $E_{n+1}$.
This is conditional to $q$, and hence to $\dcal$ and $\Ecal$.
% for each specific $\xtest$, the probability is only over $E_{n+1}$
% 
Formally, we define a binary classifier, $f(\vz) = \sI[s(\vz, y_{n+1}) \ge q]$ and $g(\vz) = \E_{\vepsilon_{n+1}}[f(\vz + \vepsilon_{n+1})]$ for which we have $\beta_{n+1} := g(\xtest)$. Note that $\gX$ is a convex subset of $\sR^d$ and the score $s(\cdot, y)$ is continuous everywhere, therefore our classifier is measurable \citep{massena2025efficient}. 
% For any fixed $\Dplus$, 
We can lower bound $\tilde{\beta}_{n+1} = \min_{\xtestpert \in \gB(\xtest)} g(\xtestpert)$ and therefore $g(\xtestpert)$ for the given $\xtest$ using the existing (binary) classification certificates, e.g. \citet{cohen2019certified}.
% We further (in \autoref{thrm:certificate-convex} show that our randomized smoothing bound is convex which allows us to lower bound $\E_{\Dplus}[\lowerval[\beta, \gB]]$.

% Below, we first show how to lower bound the coverage for one draw of $\Dplus$, then we show that our lower bound is convex allowing us to use \autoref{thrm:rcp1}.

\parbold{Certified lower bound $\tilde{\beta}$} A smoothing binary certificate computes the bound $\lowerval[g(\cdot), \xtest, \gB]$ regardless of the original definition of $f$ -- mechanics of the score function or model -- and only as a function of the value $\beta =  g(\xtest)$. We use the $\lowerval[\beta, \gB]$ notation, following \citet{anonymous2024robust}. For the known $\xtest$, a (pointwise) certified lower bound on $g(\xtestpert): \xtestpert \in \gB(\xtest)$ is obtained by searching for the worst measurable binary function $h:\gX\to\{0, 1\}$ in $\gH$ (set of all measurable functions) such that $h$ has the same smooth output as $f$ at $\xtest$. Formally: \looseness=-1 
\begin{align}
\label{eq:certificate-pointwise}
    \lowerval[\beta, \gB_{r}] = \min_{h \in \gH} \Pr_{\vepsilon}[h(\tilde\vt + \vepsilon) = 1] \quad \text{s.t.} \quad \Pr_{\vepsilon}[h(\vt + \vepsilon) = 1] = \E_{\vepsilon}[g(\xtest)] = \beta
\end{align}
The pair ($\vt$, $\tilde\vt$) are called canonical points. \citet{cohen2019certified, yang2020randomized} discuss this in detail. Intuitively, the optimization in \autoref{eq:certificate-pointwise} is translation (and in some cases rotation) invariant, and with the symmetries in the ball and the smoothing scheme, for any $\vx$, and $\tilde\vx$ we can use a fixed set of canonical points. For $\ell_p$ balls, and symmetric additive smoothing (including isotropic Gaussian noise) these points are one at the center, and the other at the edge (or vertex) of the ball; i.e. $\vt=[0, 0, \dots, 0]$, and $\tilde\vt=[r, 0, \dots, 0]$. For a detailed discussion also see section D.1 from \citet{anonymous2024robust}. Since the function $f$ itself is a feasible solution to \autoref{eq:certificate-pointwise}, it is a valid lower bound for $g(\vx_{n+1})$.

% (just using the randomly shifted scores), then we tune to a coverage level to $1 - \alpha' \ge 1 - \alpha$ such that the worst coverage remains above $1 - \alpha$ over noisy inputs. While in other approached the certified bound $\lowerval$ (or $\upperval$) for each score (or for the threshold) here we apply it to the coverage guarantee $\lowerval[1 - \alpha', \gB] \ge 1 - \alpha$. TODO: Later

\begin{figure}[b]
    \input{figures/coverage-distribution.pgf}
    \caption{
        [Left] Samples from the $\mathrm{Beta}$ distribution of clean coverage, and worst-case coverage. The dashed line is the empirical average, and the overlapping solid line is $\lowerval[1 - \alpha, \gB]$. [Middle] The empirical, and theoretical worst-case coverage. [Right] The robust $1 - \alpha'$ for two smoothing schemes.
    }
    \label{fig: cov-distribution}
\end{figure}

The mean-constrained binary certificate in \autoref{eq:certificate-pointwise} is a common problem in the randomized smoothing literature. It is efficiently solvable and in many cases has a closed form solution. For the isotopic Gaussian smoothing with $\ell_2$ (and $\ell_1$) ball the lower bound is $\tilde{\beta} = \Phi_\sigma(\Phi^{-1}_\sigma(\beta) - r)$ where $\Phi_\sigma$ is the CDF of the Gaussian distribution $\gN(0, \sigma)$ \citep{kumar2020certifying}. Using the recipe from \citet{yang2020randomized}, in \autoref{sec: Extension to All Shapes and Sizes}, we discuss how to compute $\lowerval[p, \gB]$ for other smoothing schemes.

% proposes a differentiation-based recipe to certify any other smoothing scheme. While their recipe returns a robust radius (a radius for which the lower bound probability remains above $\frac{1}{2}$), i

% \parbold{Convexity of $\lowerval$}
% if the coverage probability (continuous value in $[0, 1]$) is $\Pr_{\vepsilon}[y_{n+1} \in \gC_0(\xtest)] \ge \beta$, by moving to the worst case input $\xtestpert \in \gB(\xtest)$ the vanilla sets $\gC_0$ still obtains $\Pr_{\vepsilon}[y_{n+1} \in \gC_0(\xtestpert)] \ge \lowerval[\beta; \gB_r]$. 
% An important ingredient in \autoref{thrm:rcp1} is the following:
% As a result to bound the coverage guarantee, we need to bound $\E_{\dcal \cup \{(\vx_{n+1}, y_{n+1})\}}[\lowerval[\beta_{n+1}; \gB]]$ for $\beta_{n+1} := \Pr_{\vepsilon}[f(\vx_{n+1})\mid \dcal \cup \{(\vx_{n+1}, y_{n+1})\}]$. Intuitively we are assuming that for every draw of $\dcal\cup\{(\xtest, y_{n+1})\}$, $xtestpert$ is at its worst. We prove this property by showing that the certified lower bound problem (as in \autoref{eq:certificate-pointwise}) is convex:
\begin{lemma}
    \label{thrm:certificate-convex}
    $\lowerval[\beta, \gB]$ as the solution to \autoref{eq:certificate-pointwise} is convex and monotonically increasing w.r.t. $\beta$.
\end{lemma}
We defer the proofs to \autoref{sec:proofs}. There we rigorously prove \autoref{thrm:certificate-convex} directly from the definition of \autoref{eq:certificate-pointwise} via duality. Here we provide a sketch of an alternative proof that is insightful. \citet{lee2019tight} show that to solve \autoref{eq:certificate-pointwise}, the space $\gX$ can be divided to (finite or infinite) regions of constant likelihood ratio $\gR_t = \{ \vz: \frac{\Pr[\vz = \tilde\vt + \epsilon]}{\Pr[\vz = \vt + \epsilon]} = c_t\}$. If we can sort these regions in descending order w.r.t $c_t$, the problem reduces to the following linear program which is a fractional knapsack problem:
\[
\lowerval[\beta, \gB] = \min_{\vh \in [0, 1]^T} \vh^\top \cdot \vq \quad \text{s.t.} \quad \vh^\top \cdot \vp = \beta
\]
where $T$ is number or regions, $h_t$ is the average value of $h(\vz)$ inside the region $\gR_t$, 
$p_t = \Pr[\vt + \vepsilon \in \gR_t], q_t = \Pr[\tilde \vt + \vepsilon \in \gR_t]$, and the vectors $\vh, \vp, \vq$ gather all $h_t, p_t, q_t$'s (see \citet{lee2019tight} for the derivation).
W.l.o.g assume that the $\vp$, and $\vq$ are sorted decreasingly w.r.t. $c_t$. The optimal solution is $\vh^{\star} = [1, 1, \dots, m, 0, \dots, 0]$, for some $m \in (0, 1)$ which is to fill regions in order up to when $\vh\cdot\vp$ reaches $\beta$. 
Each index of $\vh^{\star}$ is one region being filled and by setting $h^*_t = 1$ the $\vh\cdot\vq$, and $\vh\cdot\vp$ increase by $q_t$, and $p_t$.
Therefore $\lowerval[\beta, \gB] = \vh^{\star} \cdot \vq$ is a continuous piecewise linear function with slope of $q_t / p_t$ which is increasing across regions. A piecewise linear function with an increasing slope in each piece is convex.
This convexity directly helps us to bound $\E[\lowerval[\beta, \gB]] \ge \lowerval[\E[\beta], \gB]$.

Note that in \autoref{thrm:rcp1} the guarantee is over the coverage probability and independent of the setup; therefore, without any change, one can use it to make conformal regression robust. Regardless of the downstream task the certificate is always for binary classification. Furthermore, the result is not restricted to a specific scheme and can be used for any smoothing and perturbation ball (see \autoref{sec: Extension to All Shapes and Sizes}). 

\parbold{Coverage distribution}
The expected coverage probability is itself a random variable with expectation higher than $1 - \alpha$. Under mild assumptions, $\Pr[S_{n+1} \ge q \mid \dcal] \sim \mathrm{Beta}((1-\alpha)\cdot(n+1), \alpha(n+1))$ \citep{angelopoulos2024theoretical}.
% This variable comes from a beta-distribution and for each sample $\beta$ from it, the robust coverage $\tilde\beta$ (under worst-case noise) is lower bounded by $\lowerval[\beta, \gB]$ -- note that in \autoref{sec: Robust CP with a Single Sample} we referred to $\beta$ as the coverage distribution over the noise for a single test point $\xtest$; here we refer to it as $\Pr[y_{n+1} \in \gC(\xtest) | \dcal]$ which is coming from the beta distribution. 
That is, for any given fixed calibration set, the coverage fluctuates around $1-\alpha$, with variance inversely proportional to the size of the calibration set. Since in practice we only have one calibration set, understanding this distribution, and its variance, is important.
While we do not know the distribution of the robust coverage, we can compute a conservative estimate by convolving the CDF of $\mathrm{Beta}$ and the function $\lowerval[\cdot, \gB]$ (see \subautoref{fig: cov-distribution}{left}). Similarly, convexity helps to bound the mean of this new distribution as shown in \autoref{thrm:rcp1}. Note, our method does not take into account the distribution of the scores or inputs (unlike BinCP \cite{anonymous2024robust}) and as a result it is very conservative. In \subautoref{fig: cov-distribution}{middle} we show comparison of our guaranteed lower-bound coverage and the empirical coverage under adversarial attack, highlighting that our guarantee accounts for significantly more damage. 
% \subautoref{fig: cov-distribution}{middle} compares the worst case guarantee and the empirical performance of \method (with $\gC_0$) under adversarial attack.

\parbold{Maintaining $1 - \alpha$ coverage}\autoref{thrm:rcp1} says that under perturbation, the coverage guarantee of CP calibrated with $1  - \alpha$ over augmented inference decreases at most by $\lowerval[1 - \alpha, \gB]$. A simple solution to attain $1 - \alpha$ robust coverage is to set the nominal coverage to a value $1 - \alpha'$ such that $\lowerval[1 - \alpha', \gB] \ge 1 - \alpha$. In general, we can find $1 - \alpha'$ using binary search, however, from \citep{anonymous2024robust} we know that in smoothing schemes like Gaussian, we have $\upperval[\lowerval[p, \gB], \gB^{-1}] = p$ (see \autoref{sec:choosing-conservative}, \autoref{thrm:phighlow}). Therefore, to attain $1 - \alpha$ robust coverage, we only need to set the threshold as the $\upperval[1 - \alpha, \gB^{-1}]$ quantile of the calibration scores (see \subautoref{fig: cov-distribution}{right}).

% \begin{figure}

%     \centering
%     \includegraphics{illusterative.pdf}
%     \caption{Caption}
%     \label{fig:placeholder}
% \end{figure}

% \begin{algorithm}[H]
% \caption{Robust conformal prediction with one sample (\method); the \textcolor{teal}{colored} text shows what is different from the conventional CP.}
% \label{alg:split-conformal}
% \begin{algorithmic}[1]
% \Require Calibration set $\mathcal{D}=\{(\vx_i,y_i)\}_{i=1}^n$; nominal coverage $1-\alpha\in(0,1)$; score function $s:\mathcal{X}\times\mathcal{Y}\to\mathbb{R}$; test point $\xtestpert$
% \State Compute calibration scores $s_i \gets s\!\left(\vx_i \textcolor{teal}{+ \vepsilon},\,y_i\right)$ for $(\vx_i, y_i)\in\dcal$.
% \State \textcolor{teal}{Find} $\textcolor{teal}{1 - \alpha' \gets \upperval[1 - \alpha, \gB^{-1}]}$ \Comment{e.g. for Gaussian smoothing it is $\Phi_\sigma(\Phi_\sigma^{-1}(1 - \alpha) + r)$.}
% \State Set $\bar{q}_\alpha = \quantile{\textcolor{teal}{\alpha'}}{\{s_i\}_{i = 1}^n}$.
% \State For input $\xtestpert$ \Return $\gC_r(\xtestpert) = \{y: s(\xtestpert \textcolor{teal}{+ \vepsilon}, y) \ge \textcolor{teal}{\bar{q}}\}$
% \end{algorithmic}
% \end{algorithm}

% --- Common choices of s(x,y) (optional, keep or remove) ---
% Regression (absolute residual): s(x,y) = |y - \hat f(x)|
% Regression (symmetric interval via predictive std \hat\sigma): s(x,y) = |y - \hat f(x)|/\hat\sigma(x)
% Classification (plug-in set): s(x,y) = 1 - \hat p_{\hat f}(y \mid x), so \widehat{\mathcal{C}}_\alpha(x) = { y : 1 - \hat p(y|x) \le \hat q_\alpha }
% APS/TPS can be used by redefining s(x,y) appropriately.

\subsection{Robust Conformal Sets with Randomized Smoothing of All Shapes and Sizes}
\label{sec: Extension to All Shapes and Sizes}

Both \method, and BinCP work with any smoothing and ball $\gB$. However, some binary certificates are given as a robust radius $r^*$ -- the radius up to which the prediction remains the same, i.e. $\lowerval[p, \gB_{r^*}] = 0.5$. \citet{yang2020randomized} provide a recipe to compute the $r^*$ for general $\ell_p$ certificate under additive randomized smoothing. We tweak their ``differential'' method to derive probability bounds.
We phrase \autoref{thrm:allshapes} in a notation close to \cite{yang2020randomized} and far from our own, however the takeaway is simple: in short we define $\Omega(p)$ such that $1/\Omega(p)$ encodes the minimum perturbation to make an infinitesimal increase in the worst case classifier with expected value $\beta$. We use line integral to find the $\upperval$ for the worst case classifier at radius $r$; formally $\sup\{\overline{\beta}: \int_{\beta}^{\overline{\beta}}\frac{1}{\Omega(p)}\differ p \le r\}$. We can find this supremum either analytically (see \autoref{sec: Lower and Upper Bounds for All Shapes and Sizes}) or via binary search using the existing closed forms provided by \citet{yang2020randomized}.\looseness=-1

\begin{proposition}
\label{thrm:allshapes}
    For a binary classifier $f(\vx)$, and an additive smoothing function $\xi$, 
    % let $g(\vx):=\Pr_{\vepsilon\sim \xi}[f(\vx + \vepsilon) = 1]$. 
    Let $\gU = \{\vx: f(\vx) = 1\}$ be the decision boundary and $\gU - \vz$ as the same set translated by $-\vz$, such that $\vz$ goes to the origin. Let $\xi(\gU) = \Pr_{\vepsilon \sim \xi}[\vepsilon \in \gU]$ be the expectation of the decision boundary under smoothing, and $\beta = \xi(\gU - \vx)= \E[f(\vx + \vepsilon)] $. Define: \[
    \Omega(p) := \sup_{\vdelta: \|\vdelta\| = 1} \quad \sup_{\gU \in \sR^d: \xi(\gU) = p} \quad \lim_{r \to 0^+} \frac{\xi(\gU - r\vdelta) - p}{r}
    \]
    Assuming $\Omega(p)$ is strictly positive for $p \in [\lowerval[\beta, \gB_r], \upperval[\beta, \gB_r]]$, and defining $F(\gamma) = \int_{\gamma}^{1/2} \frac{1}{\Omega(p)}\differ p$:
    % , for all $r \le F(\beta)$:
    \begin{align}
        \label{eq:all-shapes-upper}
        \upperval[\beta, \gB_r] = \sup\left\{ \beta': F(\beta') \ge F(\beta) - r \right\}
    \end{align} 
    Similarly, we have $
        \lowerval[\beta, \gB_r] = \inf\left\{\beta': F(\beta') \le F( \beta) + r \right\}
    $.
\end{proposition}

% The function $1 / \Omega(p)$ encodes the minimum perturbation to make an infinitesimal increase in the expectation of the worst-case classifier \citep{yang2020randomized}. Intuitively, the differential approach uses the line integral to bound the changes in the function $g(\vx)$ when moving to $g(\tilde\vx)$.
% 
% One way to obtain the bound is compute the closed form of $\Omega(p)$ for the smoothing and find $\overline{\beta}$ in  $\int_{\beta}^{\overline{\beta}}\frac{1}{\Omega(p)}\differ p \le r$. \citet{yang2020randomized} derive the closed form of $\Omega(p)$ (originally written as $\Phi(p)$) for various distributions. We can also directly use their result $F(\beta)$, and find the upper bound via binary search.
% -- $F(\beta)$ is increasing for $\beta \in [0, 0.5]$.

\subsection{Extension to Conformal Risk Control}
We use robustness certificates to define smoothing-based robust risk control for the first time. Let $\gC_\lambda(\cdot)$ be a conformal set, where $\lambda \leq \lambda_\mathrm{max}$ controls the set size.
For a risk function $\gL(\vx_i, y_i; \lambda) \in [a, b]$ that is right-continuous and non-increasing w.r.t. $\lambda$, if $\gL(\vx_i,y_i;  \lambda_\mathrm{max}) \le \alpha$, \citet{angelopoulos2022conformal} show: \[
\E_{\gD_{n+1}}[\gL(\vx_{n+1}, y_{n+1}; \lambda^*)] \le \alpha \quad \text{for} \quad \lambda^* = \inf\{\lambda: \frac{\sum_{i = 1}^{n}\gL(\vx_i, y_i; \lambda) + b}{n+1} \le \alpha\}
\]
Here $\alpha \in [a, b]$ is any user adjusted risk level. Similar to conformal prediction, we can also define a randomly augmented risk function $\gL(\vx_i + \vepsilon_i, y_i;  \lambda)$. The noise does not break the exchangeability and therefore $\E[\gL(\vx_{n+1} + \vepsilon_{n+1}, y_{n+1}; \lambda^*)] \le \alpha$ for the $\lambda^*$ computed on the randomly augmented calibration set. Due to the continuous nature of the risk function, we now use confidence certificates: \begin{align}
    \label{eq:risk-certificate}
    \upperval_{c}[\beta, \gB] = \max_{h\in \gH} \E[h(\xtestpert)] \quad \text{s.t.} \quad \E[h(\xtest)] \le \beta
\end{align}
Here $h:\gX\to[0, 1]$.
Similarly, \autoref{eq:risk-certificate} can be efficiently solved, and for the Gaussian distribution it has a closed form solution of $b\cdot\Phi_\sigma(\Phi_\sigma^{-1}(\frac{\beta - a}{b - a}) + r) - a(1 - \Phi_\sigma(\Phi_\sigma^{-1}(\frac{\beta - a}{b - a}) + r))$. With $[a, b] = [0, 1]$ (e.g. for the false negative rate risk) the closed form is identical to the classification certificate \citep{kumar2020certifying}.

\section{Experiments}
\label{sec: Experiments}

\parbold{Metrics and Baseline} We evaluate average set size (lower is better), and empirical coverage (exceeding $1 - \alpha$ on average). Note that in RCPs the empirical coverage conservatively exceeds $1 - \alpha$ by increasing $r$. Under perturbation this decreases at worst to $1 - \alpha$. As BinCP  \citep{anonymous2024robust} outperforms other robust CP approaches \citep{zargarbashirobust,yan2024provably}, we set it as our main comparison baseline. 
% Originally, BinCP is designed to decrease the required sample rate for efficient (small) robust sets. 
All recent smoothing-based RCPs return non-informative sets ($\gC(\vx) = \gY$) for low number of samples (e.g. $\le 32$). Note that our main contribution is to return efficient sets with \emph{one inference per input}; therefore we do not expect \method to outperform BinCP for a large sample-rate. Our reported results are over 100 iterations with calibration set randomly sampled from the data. 
 Further details are in \autoref{sec: Supplementary Experiments}, and the code is in our \href{https://github.com/soroushzargar/RCP1}{GitHub}.\looseness=-1

Since we certify the coverage guarantee (instead of scores), we can use the same \emph{binary} certificate for both classification and regression tasks. We discuss the classification here, and defer the regression task to \autoref{sec: Supplementary Experiments}. The algorithm remains the same, only for the regression we use the absolute distance from the ground truth as the score. To the best of our knowledge, this is the first conformal regression certificate based on randomized smoothing.

\parbold{Classification}
We compare methods for the CIFAR10, and ImageNet datasets. We have two inference pipelines 
% \begin{enumerate*}[label=(\roman*)]
    % \item
    The original pipeline from BinCP, and CAS (computationally cheap setup): we use the ResNet models trained with noise augmentation from \citet{cohen2019certified}. Because of the model size, large sample-rates, although inefficient, are not unrealistic.
    % \item 
    We also evaluate on an alternative more expensive pipeline outlined by \citet{carlini2022certified}: the input is first denoised by a diffusion model and then classified by a vision transformer. For CIFAR-10 we combine a 50M-parameter diffusion model from \citet{dhariwal2021diffusion}, with a \texttt{ViT-B/16} from \citet{dosovitskiy2020image}, pretrained on ImageNet at  $224 \times 224$ resolution and finetuned on CIFAR10 with 97.9\% accuracy for the HuggingFace implementation. For ImageNet we use a 552M-parameter class-unconditional diffusion model followed by BEiT-L model (305M parameters) from \citet{bao2021beit} achieving 88.6\% top-1 validation accuracy. We use the implementation provided by the timm library \citep{rw2019timm}.
% \end{enumerate*}

\begin{figure}[t]
    \input{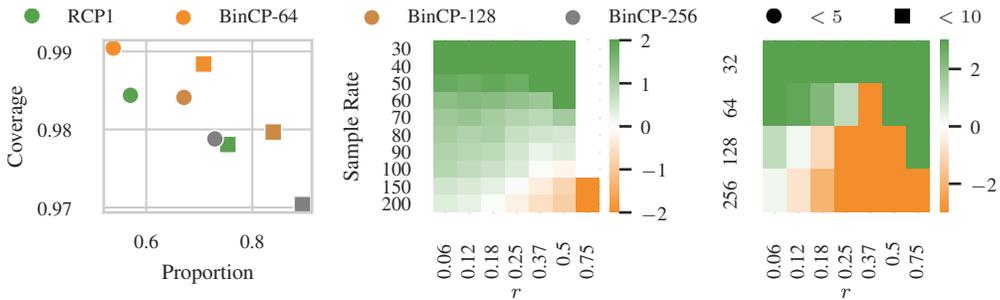}
    \caption{[Left] Proportion, and coverage of the prediction sets with $\le5$, and $\le10$ elements for the ViT model. [Middle] $|\gC_{r, \mathrm{BinCP}}| - |\gC_{r, \mathrm{\method}}|$ for the CIFAR-10 dataset with a ResNet model. [Right] ImageNet dataset and ViT models ($r=0.25$).  In all plots $\sigma=0.5$ and \method uses a single sample.}
    \label{fig:imagenet-compl}
\end{figure}

\begin{figure}[t]
    \input{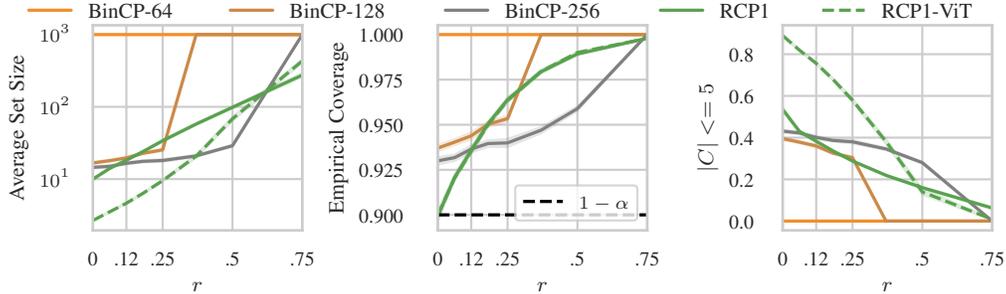}
    \caption{On ImageNet, cheap setup with $\sigma=0.5$: [Left] Compares the average set size of BinCP with various sample rates to \method. [Middle] the empirical coverage. [Right] proportion of prediction sets with size less than 5 elements. Dashed line is the result of the ViT model with the same $\sigma$.}
    \label{fig:imagenet-setsize}
\end{figure}

\begin{figure}
    \input{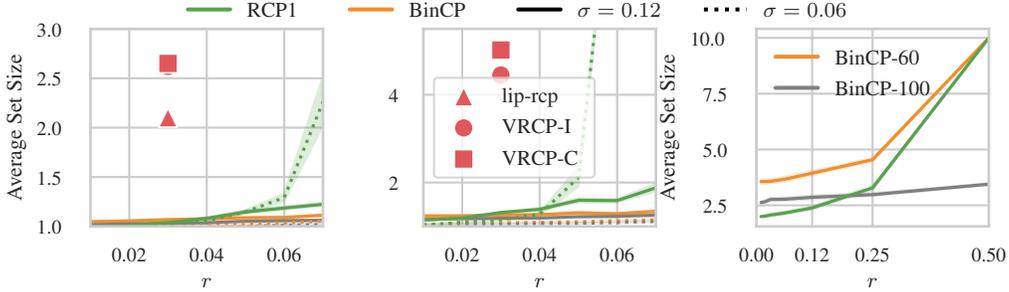}
    \caption{Performance on smaller radii in comparison with non-smoothing RCPs \citep{jeary2024verifiably,massena2025efficient} for CIFAR-10 and a ViT model. [Left] $1 - \alpha = 0.9$, and [Middle] $1 - \alpha=0.95$. Smoothing is better. [Right] Performance of the methods for a Uniform-$\ell_1$ certificate for $\epsilon \sim \mathrm{Uniform}[-1/\sqrt{3}, 1/\sqrt{3}]$.}
    \label{fig:small-r}
\end{figure}

\begin{figure}
    \input{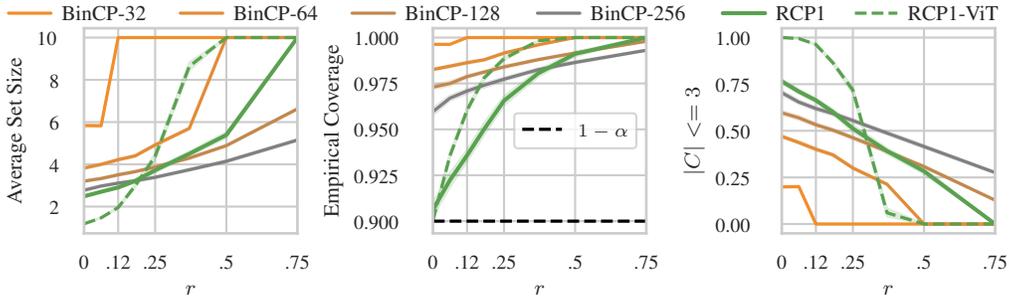}
    \caption{On CIFAR-10, ResNet model with $\sigma=0.5$: [Left] The average set size compared to BinCP (with various sample rates), \method [Middle] the empirical coverage,
    % (lower is less unnecessarily conservative),
    and [Right] proportion of sets with less than 3 labels. Dashed line shows the ViT model (with $\sigma=0.25$).}
    \label{fig:cifar-10-setsize}
\end{figure}

\parbold{Smaller set sizes} Increasing the sample-rate (number of model forwards) in BinCP decreases the set size. \subautoref{fig:first-fig}{middle} compares the set size and computation time for BinCP and \method on the ImageNet dataset. Here, \method shows similar set size to BinCP with 64 to 128 inferences per point. We also compare set size per radii for CIFAR-10 in \autoref{fig:cifar-10-setsize}, and for ImageNet in \autoref{fig:imagenet-setsize}. 
% Given that the larger pipeline (diffusion and ViT) takes considerably less time compared to the ResNet model with enough sample rate, with a small sacrifice in time, we can achieve considerably better set sizes. 
A single inference over the larger pipeline (diffusion and ViT) for \method takes significantly less time compared to the cheaper pipeline with enough samples for BinCP. Therefore we can easily achieve a considerably better set size with an unnoticeably more computation only by using a better model. 
In \subautoref{fig:imagenet-compl}{middle, and right} we compare BinCP and \method in set size per sample rate (for BinCP) and radius for the CIFAR-10 and ImageNet datasets. Our complete comparison on this experiment is in \autoref{sec: Supplementary Experiments}. Note that it is significantly inefficient to run $\ge 100$ forwards passes per image on the ViT models. Additionally, we use the results in \autoref{sec: Extension to All Shapes and Sizes} to show that the method works similarly for any smoothing scheme and threat model. For that we show the performance of BinCP (under two sample rates) and \method for the $\ell_1$ ball under uniform smoothing distribution in \subautoref{fig:small-r}{right}.

For a dataset like ImageNet (with 1000 classes), the average set size alone is not a measure of usability. Consider a CP returning $50\%$ singleton sets and $|\gY|$ for the rest, compared to a CP returning sets of size $100$ for all inputs. Surely, the latter option is not usable even though it has smaller average set size. Hence, we also report the proportion of the prediction sets with less than 5 elements in \subautoref{fig:imagenet-setsize}{right} (also see \autoref{sec: Supplementary Experiments}). This metric is only trustworthy if the we don't sacrifice the coverage in smaller sets. In \subautoref{fig:imagenet-compl}{left} we show that these sets have coverage larger than $1 - \alpha$. \looseness=-1

\begin{table}[b]
\centering
\caption{Estimated runtimes (in \texttt{HH:MM:SS}) for 1000 inputs using an H-100 GPU. Results are scaled from a full experimental run assuming a linear cost in both the number of inputs and samples.}
\vspace{0.5em}

\begin{tabular}{@{}llcccc@{}}
\toprule
\rowcolor{gray!10}
\textbf{Pipeline} & \textbf{Dataset} & \textbf{1 Sample} & \textbf{64 Samples} & \textbf{128 Samples} & \textbf{256 Samples} \\
\midrule
\multirow{2}{*}{ViT} 
 & CIFAR-10  & 0:00:01 & 0:01:09 & 0:02:19 & 0:04:39 \\
 & ImageNet  & 0:00:33 & 0:35:30 & 1:11:00 & 2:22:01 \\
\bottomrule
\end{tabular}

\vspace{-0.5em}
\label{tab:smoothing_runtime}
\end{table}

\begin{table}[b]
\centering
\caption{Risk and mask size for the Cityscapes dataset. Risk level is $0.15$, with 100 calibration points. The variance is not over calibration sampling but over the images and $r=0.06$.}
\vspace{0.5em}
\begin{adjustbox}{width=0.95\linewidth}
\begin{tabular}{@{}lccccc@{}}
\toprule
\rowcolor{gray!10}
\textbf{Class} & \textbf{Risk} & \textbf{Robust Risk} & \textbf{(True) Class Prop.} & \textbf{Mask Prop.} & \textbf{Robust Mask Prop.} \\
\midrule
Pedestrian & 0.1474 $\pm$ 0.2797 & 0.1111 $\pm$ 0.2588 & 0.0160 $\pm$ 0.0279 & 0.1891 $\pm$ 0.1257 & 0.2522 $\pm$ 0.1312 \\
Car & 0.1466 $\pm$ 0.2582 & 0.0833 $\pm$ 0.2032 & 0.0539 $\pm$ 0.0545 & 0.0832 $\pm$ 0.0733 & 0.1101 $\pm$ 0.0807 \\
\bottomrule
\end{tabular}
\end{adjustbox}
\vspace{-0.5em}
\label{tab:risk}
\end{table}

\parbold{Small radii} \citet{jeary2024verifiably}, and \citet{massena2025efficient} propose RCP using verifiers and Lipschitz constant of the network. Although their result is for one order of magnitude smaller radii (e.g. 0.02 instead of 0.25), their methods are efficient by using one forward per input. With \method being the same in that metric, we compare with them in \subautoref{fig:small-r}{left and middle} reporting performance on smaller radii. Aside better performance, \method has a black-box access and works for any model. 
Intuitively, as shown in \subautoref{fig:first-fig}{left}, smooth (or augmented) inference is significantly more robust to perturbations.

\parbold{Time-comparison} 
With $t_\mathrm{cert}$, and $t_{f}$ as the time for computing bounds, and for the model's inference time, other smoothing based RCPs at best require $\gO(n_\mathrm{mc} \times \dcal \times t_f + t_\mathrm{cert})$ for calibration where $n_\mathrm{mc}$ is the number of MC samples. For each test point they also require $\gO(n_\mathrm{mc} \times t_f)$ time. \method takes the same time as the normal model's inference plus an additional $\gO(t_\mathrm{cert})$ for calibration. Similarly, \method takes $n_\mathrm{mc}$ less memory compared to other smoothing RCPs.
We show the runtime of the ViT pipeline for the used datasets in \autoref{tab:smoothing_runtime}. Note that this is only the time to compute logits as the other processes (including certificates) are negligiable compared to it. The runtime of BinCP with a sample rate comparable to \method is significantly high for large models like ViT; for instance, \method and a comparable BinCP (with 128 samples on ImageNet) need $\sim$2',46", and $\sim$5h 55' to process 5000 images. \looseness=-1

%%%%%%%%%%%%%%% HERE for 3

\begin{figure}
    \centering
    \input{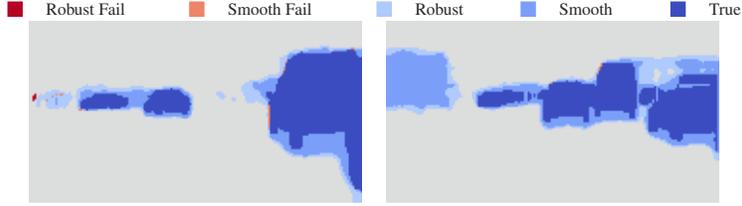}
    \caption{Performance of vanilla and robust risk control. From lighter to darker the colors are robust region, vanilla (non-robust) region, and the ground truth region. Here $\sigma=0.25$, and $r=0.06$.}
    \label{fig:risk}
\end{figure}

\parbold{Robust conformal risk control}
We use the model from \citet{fischer2021scalable} on the CityScapes dataset \citep{Cordts2016Cityscapes} which is a scene segmentation task. We mask the regions where a target class (e.g. car) might be present. The error function is the false negative ratio (FNR) -- the portion of the pixels from the target class that is not masked. We take the $\exp(f(\vx))_y$ as the score of the class $y$, and we set the mask as $\sI[\exp(f(\vx))_y \ge 1 - \lambda]$.  Note that here the classes could possibly overlap. We calibrate by finding a $\lambda$ that results in a FNR loss less than the user adjusted tolerable risk. So far, this is the first result for smoothing-based robust conformal risk control. 
% The process of robust risk control is similar to robust CP.
Similar to \method, we first smooth the image data (one sample), then we compute the $\lambda$ that results in $\lowerval_c[\alpha, \gB]$ risk. We report the results in \autoref{tab:risk}, and show an example in \autoref{fig:risk}. \looseness=-1

\begin{figure}
    \centering
    \input{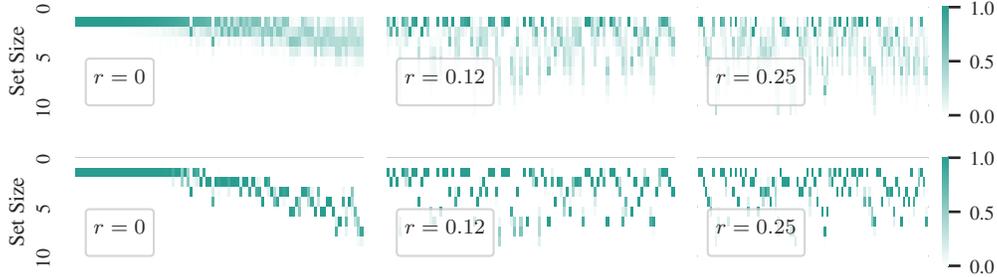}
    \caption{Randomness in set sizes for a CIFAR-10 dataset and ResNet model. The $x$-axis sorts datapoints in a fixed order (same across all plots), and the $y$ axis shows the set size. The intensity of pixels shows the probability of a specific set size for that point  over $\vepsilon$.  [Top] shows \method, and [Bottom] BinCP (150 samples). \method has larger variance compared to BinCP.}
    \label{fig:randomness}
\end{figure}

\parbold{Limitation: Increased variance} In \autoref{fig:randomness}, \method shows considerably more randomness in the prediction sets compared to BinCP. This is essentially due to the random definition of the prediction set and the score function -- the prediction set in \method is a function of the random variable $\vepsilon$. 
This randomness does not affect the final robust / vanilla coverage.

\section{Conclusion}
While offering small sets for larger radii, smoothing-based RCP methods 
% can return small prediction sets robust to large radii, however, their limitation is that they 
need many forward passes per input. Instead, we show that noise-augmented inference combined with CP is inherently robust, and with that, we propose \method which needs only one forward pass per input. Our approach returns sets with size similar to state of the art while nullifying the need of many MC samples. 
% We further extend our method to conformal risk control. 
Prior smoothing RCPs provide their guarantee by lower bounding the scores, hence they need to estimate (some statistic about) the distribution of score for each individual input. Alternatively we only apply the lower bound on the coverage guarantee which is known in prior to be $1 - \alpha$.
\newpage

\section*{Acknowledgements}
We thank Guiliana Thomanek, and Jimin Cao for their feedback on our paper.

% \section*{References}
\bibliographystyle{plainnat}
\bibliography{paper}

@article{zhang2020black,
  title={Black-box certification with randomized smoothing: A functional optimization based framework},
  author={Zhang, Dinghuai and Ye, Mao and Gong, Chengyue and Zhu, Zhanxing and Liu, Qiang},
  journal={Advances in Neural Information Processing Systems},
  volume={33},
  pages={2316--2326},
  year={2020}
}

@article{jeary2024verifiably,
      title={Verifiably Robust Conformal Prediction}, 
      author={Linus Jeary and Tom Kuipers and Mehran Hosseini and Nicola Paoletti},
      year={2024},
      eprint={2405.18942},
      archivePrefix={arXiv},
}

@article{angelopoulos2024theoretical,
  title={Theoretical foundations of conformal prediction},
  author={Angelopoulos, Anastasios N and Barber, Rina Foygel and Bates, Stephen},
  journal={arXiv preprint arXiv:2411.11824},
  year={2024}
}

@article{dhariwal2021diffusion,
  title={Diffusion models beat gans on image synthesis},
  author={Dhariwal, Prafulla and Nichol, Alexander},
  journal={Advances in neural information processing systems},
  volume={34},
  pages={8780--8794},
  year={2021}
}

@article{bao2021beit,
  title={Beit: Bert pre-training of image transformers},
  author={Bao, Hangbo and Dong, Li and Piao, Songhao and Wei, Furu},
  journal={arXiv preprint arXiv:2106.08254},
  year={2021}
}

@article{dosovitskiy2020image,
  title={An image is worth 16x16 words: Transformers for image recognition at scale},
  author={Dosovitskiy, Alexey and Beyer, Lucas and Kolesnikov, Alexander and Weissenborn, Dirk and Zhai, Xiaohua and Unterthiner, Thomas and Dehghani, Mostafa and Minderer, Matthias and Heigold, Georg and Gelly, Sylvain and others},
  journal={arXiv preprint arXiv:2010.11929},
  year={2020}
}

@misc{rw2019timm,
  author = {Ross Wightman},
  title = {PyTorch Image Models},
  year = {2019},
  publisher = {GitHub},
  journal = {GitHub repository},
  doi = {10.5281/zenodo.4414861},
  howpublished = {\url{https://github.com/rwightman/pytorch-image-models}}
}

@inproceedings{
    anonymous2024robust,
      title={Robust Conformal Prediction with a Single Binary Certificate},
      author={Zargarbashi, Soroush H and Bojchevski, Aleksandar},
      journal={arXiv preprint arXiv:2503.05239},
      year={2025}
    }

@article{carlini2022certified,
  title={(Certified!!) adversarial robustness for free!},
  author={Carlini, Nicholas and Tramer, Florian and Dvijotham, Krishnamurthy Dj and Rice, Leslie and Sun, Mingjie and Kolter, J Zico},
  journal={arXiv preprint arXiv:2206.10550},
  year={2022}
}

@inproceedings{fischer2021scalable,
  title={Scalable certified segmentation via randomized smoothing},
  author={Fischer, Marc and Baader, Maximilian and Vechev, Martin},
  booktitle={International Conference on Machine Learning},
  pages={3340--3351},
  year={2021},
  organization={PMLR}
}

@article{Romano2020ClassificationWV,
  title={Classification with Valid and Adaptive Coverage},
  author={Yaniv Romano and Matteo Sesia and Emmanuel J. Cand{\`e}s},
  journal={arXiv: Methodology},
  year={2020}
}

@inproceedings{zargarbashirobust,
  title={Robust Yet Efficient Conformal Prediction Sets},
  author={Zargarbashi, Soroush H and Akhondzadeh, Mohammad Sadegh and Bojchevski, Aleksandar},
    year={2024},
  booktitle={Forty-first International Conference on Machine Learning}
}

@article{lee2019tight,
  title={Tight certificates of adversarial robustness for randomly smoothed classifiers},
  author={Lee, Guang-He and Yuan, Yang and Chang, Shiyu and Jaakkola, Tommi},
  journal={Advances in Neural Information Processing Systems},
  volume={32},
  year={2019}
}

@inproceedings{gendler2021adversarially,
  title={Adversarially robust conformal prediction},
  author={Gendler, Asaf and Weng, Tsui-Wei and Daniel, Luca and Romano, Yaniv},
  booktitle={International Conference on Learning Representations},
  year={2021}
}

@inproceedings{Vovk2005AlgorithmicLI,
  title={Algorithmic Learning in a Random World},
  author={Vladimir Vovk and Alexander Gammerman and Glenn Shafer},
  year={2005}
}

@inproceedings{Guo2017OnCO,
  title={On Calibration of Modern Neural Networks},
  author={Chuan Guo and Geoff Pleiss and Yu Sun and Kilian Q. Weinberger},
  booktitle={International Conference on Machine Learning},
  year={2017},
  url={https://api.semanticscholar.org/CorpusID:28671436}
}

@inproceedings{Ghosh2023ProbabilisticallyRC,
  title={Probabilistically robust conformal prediction},
  author={Subhankar Ghosh and Yuanjie Shi and Taha Belkhouja and Yan Yan and Janardhan Rao Doppa and Brian Jones},
  booktitle={Conference on Uncertainty in Artificial Intelligence},
  year={2023},
  url={https://api.semanticscholar.org/CorpusID:260334753}
}

@article{yan2024provably,
  title={Provably robust conformal prediction with improved efficiency},
  author={Yan, Ge and Romano, Yaniv and Weng, Tsui-Wei},
  journal={arXiv preprint arXiv:2404.19651},
  year={2024}
}

@inproceedings{yang2020randomized,
  title={Randomized smoothing of all shapes and sizes},
  author={Yang, Greg and Duan, Tony and Hu, J Edward and Salman, Hadi and Razenshteyn, Ilya and Li, Jerry},
  booktitle={International Conference on Machine Learning},
  pages={10693--10705},
  year={2020},
  organization={PMLR}
}

@inproceedings{cohen2019certified,
  title={Certified adversarial robustness via randomized smoothing},
  author={Cohen, Jeremy and Rosenfeld, Elan and Kolter, Zico},
  booktitle={international conference on machine learning},
  pages={1310--1320},
  year={2019},
  organization={PMLR}
}

@article{kumar2020certifying,
  title={Certifying confidence via randomized smoothing},
  author={Kumar, Aounon and Levine, Alexander and Feizi, Soheil and Goldstein, Tom},
  journal={Advances in Neural Information Processing Systems},
  volume={33},
  pages={5165--5177},
  year={2020}
}

@article{massena2025efficient,
  title={Efficient Robust Conformal Prediction via Lipschitz-Bounded Networks},
  author={Massena, Thomas and And{\'e}ol, L{\'e}o and Boissin, Thibaut and Friedrich, Corentin and Mamalet, Franck and Serrurier, Mathieu and Gerchinovitz, S{\'e}bastien},
  year={2025}
}

@inproceedings{Cordts2016Cityscapes,
title={The Cityscapes Dataset for Semantic Urban Scene Understanding},
author={Cordts, Marius and Omran, Mohamed and Ramos, Sebastian and Rehfeld, Timo and Enzweiler, Markus and Benenson, Rodrigo and Franke, Uwe and Roth, Stefan and Schiele, Bernt},
booktitle={Proc. of the IEEE Conference on Computer Vision and Pattern Recognition (CVPR)},
year={2016}
}

@article{angelopoulos2022conformal,
  title={Conformal risk control},
  author={Angelopoulos, Anastasios N and Bates, Stephen and Fisch, Adam and Lei, Lihua and Schuster, Tal},
  journal={arXiv preprint arXiv:2208.02814},
  year={2022}
}

@inproceedings{bojchevski2020efficient,
  title={Efficient robustness certificates for discrete data: Sparsity-aware randomized smoothing for graphs, images and more},
  author={Bojchevski, Aleksandar and Gasteiger, Johannes and G{\"u}nnemann, Stephan},
  booktitle={International Conference on Machine Learning},
  pages={1003--1013},
  year={2020},
  organization={PMLR}
}

@article{salman2019provably,
  title={Provably robust deep learning via adversarially trained smoothed classifiers},
  author={Salman, Hadi and Li, Jerry and Razenshteyn, Ilya and Zhang, Pengchuan and Zhang, Huan and Bubeck, Sebastien and Yang, Greg},
  journal={Advances in neural information processing systems},
  volume={32},
  year={2019}
}

@article{bojarski2016end,
  title={End to end learning for self-driving cars},
  author={Bojarski, Mariusz and Del Testa, Davide and Dworakowski, Daniel and Firner, Bernhard and Flepp, Beat and Goyal, Prasoon and Jackel, Lawrence D and Monfort, Mathew and Muller, Urs and Zhang, Jiakai and others},
  journal={arXiv preprint arXiv:1604.07316},
  year={2016}
}

@inproceedings{he2016deep,
  title={Deep residual learning for image recognition},
  author={He, Kaiming and Zhang, Xiangyu and Ren, Shaoqing and Sun, Jian},
  booktitle={Proceedings of the IEEE conference on computer vision and pattern recognition},
  pages={770--778},
  year={2016}
}

%%%%%%%%%%%%%%%%%%%%%%%%%%%%%%%%%%%%%%%%%%%%%%%%%%%%%%%%%%%%

\appendix

%%%%%%%%%%%%%%%%%%%%%%%%%%%%%%%%%%%%%%%%%%%%%%%%%%%%%%%%%%%%

\newpage

\section{Related Work}
\label{sec:related-works}

\citet{gendler2021adversarially} initially proposed robust CP resilient to adversarial examples (worst-case noise) without accounting for finite samples (asymptotically valid setup). \citet{yan2024provably} added finite sample correction and proposed a new score to return (set size) efficiency. Both mentioned works were using randomized smoothing and the mean of the smooth score to bound the worst case perturbations. \citet{zargarbashirobust} (CAS) proposed to use the CDF information of the smooth score -- a more restrictive constraint and therefore returned smaller prediction sets. All of these methods required unrealistically expensive setup with $10^4$ MC samples to be able to return acceptably small sets. \citet{anonymous2024robust} (BinCP) defined a quantile-based score over the distribution of the smooth scores, that allowed same set size as CAS with orders of magnitude less samples (e.g. 200 would be sufficient). In contrast with all of the methods \method works with a single augmented sample and without involving finite sample correction, which lies at the computation efficiency side of this trade-off. 

Orthogonal to randomized smoothing, \citet{jeary2024verifiably}, and \citet{massena2025efficient} use verifiers and Lipschitz continuity of the networks to bound the score function. Their robust radii are one order of magnitude smaller than smoothing RCP, but instead they do not require many forward passes per input. In \autoref{fig:small-r} we show that our approach (with the same computational efficiency) provides smaller prediction sets outperforming their results; plus, our approaches works for any black-box model. Notably all mentioned works provide robustness to the worst-case noise which is orthogonal to probabilistically robust CP \citet{Ghosh2023ProbabilisticallyRC}.

\section{Additional Description of Figures}
\label{sec:fig-manual}

\parbold{\subautoref{fig:first-fig}{left}} We report how the empirical coverage of $\gC_0$ (dashed lines) breaks for smooth prediction in BinCP and randomized augmented score in \method, compared to the vanilla conformal prediction (red dashed line). In all cases, we calibrate over clean calibration set and for radii $r$ we return the prediction set of $\xtestpert$ which is $\xtest$ perturbed with adversarial attacks. Specifically we use the \texttt{PGDSmooth} attack from \cite{salman2019provably} for smooth methods and conventional \texttt{PGD} attack for vanilla CP. Compared to \texttt{PGD}, the \texttt{PGDSmooth} attack performs stronger for smooth scores. The solid lines are shows the empirical coverage of $\gC_r$ from BinCP and \method on the same adversarial data. The main takeaway of the figure is to show the robustness of $\gC_r$, and the inherent resilience of smooth and augmented inference to the adversarial (worst-case) noise. The result is for CIFAR-10 dataset, \texttt{ResNet} model and $\sigma=0.5$. 

\parbold{\subautoref{fig:first-fig}{middle}} We compared the robust set size of BinCP and \method; we plotted $|\gC_{r, \mathrm{BinCP}}| - |\gC_{r, \mathrm{\method}}|$ for which lower is better. The plot is for CIFAR-10 dataset, \texttt{ResNet} model and $\sigma=0.5$. 

\parbold{\subautoref{fig:first-fig}{right}} Each point is a computation of $\gC_{r}$ shown both in time and set size. All times are divided by a single inference of the \texttt{ResNet} model. Here we evaluated two forward pipelines of cheaper \texttt{ResNet} model, and more time costly diffusion + vision transformer (as discussed in \autoref{sec: Experiments}). Both axis are log-scale and the plot is for the ImageNet dataset. Here $\sigma=0.5$.

\parbold{\subautoref{fig: cov-distribution}{left}} We took samples from the Beta distribution of the coverage -- each sample is then a number $\beta$. We computed the $\lowerval[\beta, \gB]$ and draw a distribution of the new values. For the Beta distribution, we have $n = 200$, and $1 - \alpha = 0.9$. For the $\lowerval$ function we used $\sigma=0.5$, and $r = 0.25$

\parbold{\subautoref{fig: cov-distribution}{middle}} We show the empirical coverage value under the PGDSmooth attack with $\sigma=0.5$ over various radii. We use the same sigma to show the theoretical lower bound coverage.

\parbold{\subautoref{fig: cov-distribution}{right}} We reported the $\lowerval$ function for Gaussian and Laplace smoothing both with $\sigma=0.5$. The plots are not empirical.

\parbold{\subautoref{fig:small-r}{right}} Here we use the $\ell_1$ certificate from \citet{yang2020randomized}. Here the smoothing scheme is $\epsilon = \mathrm{Uniform}[-\sigma / \sqrt{3}, \sigma / \sqrt{3}]$

\section{More on Conformal Prediction}
\label{sec: More on Conformal Prediction}

Our default score function in the manuscript is TPS (threshold prediction sets) where the score function is directly set to the softmax; $s(\vx, y) = \mathrm{Softmax}_y(f(\vx))$ for the prediction model $f$. Another choice is to use the logits of the model as the conformity score $s(\vx, y) = f(\vx)_y$. Similar to BinCP we are using a single binary certificate which do not rely on bounded score function. Another conformal prediction method called as adaptive prediction sets (APS) uses the accumulated softmax up to label $y$ as the conformity score; formally $s(\vx, y) = -[\pi(\vx, y) \cdot u + \sum_{k = 1}^{|\gY|}\pi(\vx, y) \cdot \sI[\pi(\vx, y_k) > \pi(\vx, y)]]$ where $\pi(\vx, y) = \mathrm{Softmax}_y(f(\vx))$ and $u \sim \mathrm{Uniform}[0, 1]$. While this score results in larger sets, it increases adaptivity -- approximate conditional coverage.

\begin{figure}
    \input{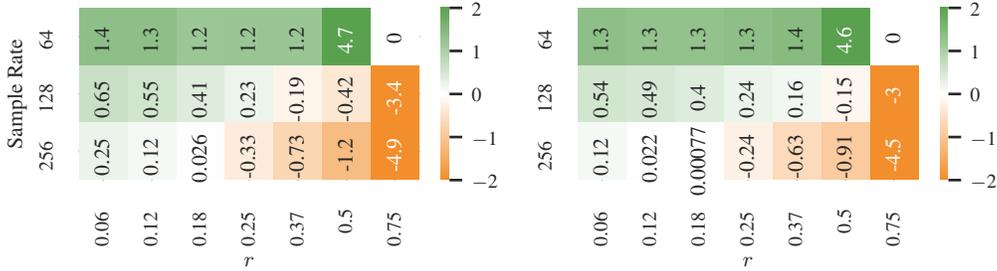}
    \caption{Comparison of BinCP and \method for [Left] TPS, and [Right] APS score function on CIFAR-10 dataset with $\sigma=0.9$.}
    \label{fig:scores-cifar}
\end{figure}

We report the result using these score functions in \autoref{fig:scores-cifar} (comparison of BinCP and \method for each score) and \autoref{tab:tps_aps_full} (the set size of \method for both score functions). As expected, similar trend as TPS is observed for APS as well.  

Same as BinCP we also do not need the score function to be bounded. However in the end, using an unbounded score function (like using logits directly) did not show to improve over the existing APS and TPS. 

\begin{table}[t]
    \centering
    \caption{Set size of \method for TPS and APS score across radii ($r$) and target coverage guarantees.}
    \vspace{0.5em}
    % \begin{adjustbox}{width=0.95\linewidth}
    
\begin{tabular}{@{}ccccccc@{}}
\toprule
\rowcolor{gray!10}
\textbf{$r$} & \textbf{Coverage} & \multicolumn{2}{c}{\textbf{TPS}} & \multicolumn{2}{c}{\textbf{APS}} \\
\cmidrule(lr){3-4} \cmidrule(lr){5-6}
& & Avg Set Size & Emp. Cov. & Avg Set Size & Emp. Cov. \\
\midrule
0.06 & 0.85 &   2.17 $\pm$ 0.02 &        0.88 $\pm$ 0.00 &   2.52 $\pm$ 0.03 &        0.88 $\pm$ 0.00 \\
     & 0.90 &   2.70 $\pm$ 0.03 &        0.92 $\pm$ 0.00 &   2.98 $\pm$ 0.02 &        0.92 $\pm$ 0.00 \\
     & 0.95 &   3.74 $\pm$ 0.01 &        0.97 $\pm$ 0.00 &   3.91 $\pm$ 0.04 &        0.97 $\pm$ 0.00 \\
0.12 & 0.85 &   2.44 $\pm$ 0.03 &        0.90 $\pm$ 0.00 &   2.76 $\pm$ 0.02 &        0.90 $\pm$ 0.00 \\
     & 0.90 &   2.93 $\pm$ 0.04 &        0.94 $\pm$ 0.00 &   3.23 $\pm$ 0.01 &        0.94 $\pm$ 0.00 \\
     & 0.95 &   3.96 $\pm$ 0.06 &        0.97 $\pm$ 0.00 &   4.07 $\pm$ 0.04 &        0.97 $\pm$ 0.00 \\
0.18 & 0.85 &   2.70 $\pm$ 0.04 &        0.92 $\pm$ 0.00 &   2.99 $\pm$ 0.01 &        0.92 $\pm$ 0.00 \\
     & 0.90 &   3.25 $\pm$ 0.05 &        0.95 $\pm$ 0.00 &   3.44 $\pm$ 0.03 &        0.95 $\pm$ 0.00 \\
     & 0.95 &   4.48 $\pm$ 0.09 &        0.98 $\pm$ 0.00 &   4.48 $\pm$ 0.02 &        0.98 $\pm$ 0.00 \\
0.25 & 0.85 &   3.03 $\pm$ 0.01 &        0.94 $\pm$ 0.00 &   3.33 $\pm$ 0.04 &        0.94 $\pm$ 0.00 \\
     & 0.90 &   3.70 $\pm$ 0.02 &        0.97 $\pm$ 0.00 &   3.89 $\pm$ 0.03 &        0.97 $\pm$ 0.00 \\
     & 0.95 &   4.81 $\pm$ 0.01 &        0.99 $\pm$ 0.00 &   4.82 $\pm$ 0.04 &        0.99 $\pm$ 0.00 \\
0.37 & 0.85 &   3.62 $\pm$ 0.06 &        0.96 $\pm$ 0.00 &   3.91 $\pm$ 0.01 &        0.97 $\pm$ 0.00 \\
     & 0.90 &   4.48 $\pm$ 0.03 &        0.98 $\pm$ 0.00 &   4.55 $\pm$ 0.09 &        0.98 $\pm$ 0.00 \\
     & 0.95 &   6.24 $\pm$ 0.14 &        0.99 $\pm$ 0.00 &   6.49 $\pm$ 0.04 &        1.00 $\pm$ 0.00 \\
0.50 & 0.85 &   4.51 $\pm$ 0.03 &        0.98 $\pm$ 0.00 &   4.55 $\pm$ 0.11 &        0.98 $\pm$ 0.00 \\
     & 0.90 &   5.32 $\pm$ 0.04 &        0.99 $\pm$ 0.00 &   5.34 $\pm$ 0.03 &        0.99 $\pm$ 0.00 \\
     & 0.95 &  10.00 $\pm$ 0.00 &        1.00 $\pm$ 0.00 &  10.00 $\pm$ 0.00 &        1.00 $\pm$ 0.00 \\
0.75 & 0.85 &   6.28 $\pm$ 0.19 &        0.99 $\pm$ 0.00 &   6.31 $\pm$ 0.11 &        0.99 $\pm$ 0.00 \\
     & 0.90 &  10.00 $\pm$ 0.00 &        1.00 $\pm$ 0.00 &  10.00 $\pm$ 0.00 &        1.00 $\pm$ 0.00 \\
     & 0.95 &  10.00 $\pm$ 0.00 &        1.00 $\pm$ 0.00 &  10.00 $\pm$ 0.00 &        1.00 $\pm$ 0.00 \\
\bottomrule
\end{tabular}

    % \end{adjustbox}
    \vspace{-0.5em}
    \label{tab:tps_aps_full}
\end{table}

\section{Supplementary to Theory}

\subsection{Robust Conformal Prediction Guarantee}
\label{sec: Threat Model}
The guarantee in \autoref{eq:robust-conformal-guarantee} doesn't take the randomness in score function and prediction set into account. This is while many conformal scores have a random variable inside, for instance APS \cite{Romano2020ClassificationWV} multiplies the probability of each class by a uniform random value to break the ties (and enable the exact $1 - \alpha$ coverage).
The original guarantee taken from \cite{Ghosh2023ProbabilisticallyRC, zargarbashirobust} is:
\begin{align*}
    \Pr_{\substack{\dcal, \xtest\sim\gD}}[y_{n+1} \in \gC_\gB(\xtestpert), \forall \xtestpert \in \gB(\xtest)] \ge 1 - \alpha \\
    \equiv \Pr_{\substack{\dcal, \xtest\sim\gD}}[\inf_{\forall \xtestpert \in \gB(\xtest)}\sI[y_{n+1} \in \gC_\gB(\xtestpert)] = 1 ] \ge 1 - \alpha
\end{align*}
This formulation fails to capture the stochasticity in the score function (and hence in the prediction set). Since with a very small probability to miscover a point (which is non-zero in methods like APS) the indicator evaluates to false. The worst-case indicator $\inf_{\forall \xtestpert \in \gB(\xtest)} \sI[y_{n+1} \in \gC_\gB(\xtestpert) = 1 ]$ becomes zero whenever there exists any probability that $\xtest$ is not covered. Consider the non-robust case where no perturbations are applied; i.e., evaluating the coverage guarantee of standard conformal prediction. Using the APS score function and shrinking the perturbation space to an infinitesimal ball $\gB_r$ as $r \to 0^+$ (and therefore $\xtestpert \to \xtest$), the coverage probability should exceed $1 - \alpha$, since APS already satisfies this guarantee. However many of the datapoints that have a small probability to exclude the true class from the prediction set will not pass the worst-case indicator. Consider a datapoint with one hot conditional probability, still the top label will be in the prediction set with probability $-q$ where $q$ is the threshold. 

While the shortcoming in \autoref{eq:robust-conformal-guarantee} excludes any CP with randomness in the score, still previous smoothing-based RCP methods (at least in assymptotically valid setup) satisfy its conditions independent of the score function used. This is since all previous methods systematically remove all the randomness from the score and return a deterministic prediction set. Given any base score function, these methods defined their own score as a statistic (e.g. mean, or quantile) over the distribution of the base score on $\vx + \vepsilon$. As the distribution already includes the inherent randomness in the base score itself, the statistics like mean \cite{gendler2021adversarially, zargarbashirobust} and the quantile \cite{anonymous2024robust} are deterministic (excluding any randomness). This is an orthogonal to the probabilistic nature of estimating these statistics from Monte Carlo samples (the validity of confidence intervals). As a result the final set based on these scores exclude the inherent randomness of the base score function.

\subsection{Proofs}
\label{sec:proofs}
\begin{proof}[Proof of \autoref{thrm:certificate-convex}]
    The Lagrangian form of \autoref{eq:certificate-pointwise} is:
    \begin{align*}
        \gL(\beta, \lambda) = \min_{h\in \{0, 1\}^{|\gX|}} \Pr_{\vepsilon}[h(\tilde\vt + \vepsilon) = 1] - \lambda\left( \Pr_{\vepsilon}[h(\vt + \vepsilon) = 1] - \beta \right) \\
        = \min_{h\in \{0, 1\}^{|\gX|}} E_{\vz \sim q}[h(\vz)] - \lambda\cdot E_{\vz \sim p}[h(\vz) - \beta] = \lambda \cdot \beta + 
        \min_{h\in \{0, 1\}^{|\gX|}} E_{\vz \sim q}[h(\vz)] - E_{\vz \sim p}[h(\vz)] \\
        = \lambda \cdot \beta + \min_{h\in \{0, 1\}^{|\gX|}} \int_\gX(q(\vz) - \lambda \cdot p(\vz))\cdot h(\vz)\mathrm{d}\vz 
    \end{align*}
    where $p$ and $q$ are smoothing distributions centered at $\vt$ and $\tilde \vt$ respectively.
    The worst classifier (the minimizer of the problem) can be derived as follows
    \begin{align*}
        h(\vz) = \begin{cases}
            0 & \text{if} \quad q(\vz) - \lambda \cdot p(\vz) \ge 0\\
            1 & \text{otherwise}
        \end{cases}        
    \end{align*}
    Intuitively, to minimize the term $\int_\gX(q(\vz) - \lambda \cdot p(\vz))\cdot h(\vz)\mathrm{d}\vz$ we look at each point independently. For each point if the term $(q(\vz) - \lambda \cdot p(\vz))$ is positive we cancel it by $h=0$ and if negative we keep it to decrease the total integral value. The resulting dual with the dual variable $\lambda \ge 0$ is then:
    \begin{align*}
        \gL(\beta, \lambda) = \lambda \cdot \beta + \int \min\{0, p(\vz) - \lambda \cdot q(\vz) \mathrm{d}\vz\} = \lambda \cdot \beta + l(\lambda)
    \end{align*}
    Here $l(\lambda):= \int \min\{0, p(\vz) - \lambda \cdot q(\vz) \mathrm{d}\vz\}$ is only a function of $\lambda$. Maximizing over $\lambda$ we get the optimal dual soltuion which equals the optimal primal since it was shown that strong duality holds \citep{zhang2020black}:
    \begin{align*}
        \lowerval[\beta, \gB] = \max_\lambda \lambda\cdot \beta + l(\lambda)
    \end{align*} Which is pointwise maximum of affine functions and therefore convex in $\beta$.

    The monotonicity w.r.t. $\beta$ directly follows from the definition. By increasing $\beta$ the feasible space reduces to a nested subset of the previous problem which means that the solution will be greater than or equal to the original solution.
\end{proof}

\subsection{Choosing the conservative $1 - \alpha'$}
\label{sec:choosing-conservative}
In general to obtain $1 - \alpha$ robust coverage guarantee, we should choose the nominal $1 - \alpha'$ in \autoref{alg:split-conformal} such that $\lowerval[1 - \alpha, \gB] \ge 1 - \alpha$. This nominal probability can be found via binary search due to the non-decreasing nature of $\lowerval$. But in many cases including the Gaussian distribution, where the canonical points can be used interchangeably (choosing $(\vt, \tilde\vt)$ as the pair of clean, and noisy canonical points doesn't differ from the opposite $(\tilde\vt, \vt)$, the following lemma allows us to set the $1 - \alpha' = \upperval[1 - \alpha, \gB^{-1}]$.

\begin{lemma}
\label{thrm:phighlow}
    If for a smoothing scheme, and a perturbation ball $\gB$, canonical points $\vt$, and $\tilde\vt$ can be used interchangeably; then we have $\upperval[\lowerval[p, \gB], \gB^{-1}] = p$. Using the canonical points interchangeably means that for $\upperval$, and (similarly $\lowerval$) both of the optimizations
    \[
    \max_{h \in \gH} \Pr_{\vepsilon}[h(\tilde\vt + \vepsilon) = 1] \quad \text{s.t.} \quad \Pr_{\vepsilon}[h(\vt + \vepsilon) = 1] = p 
    \] and \[
    \max_{h \in \gH} \Pr_{\vepsilon}[h(\vt + \vepsilon) = 1] \quad \text{s.t.} \quad \Pr_{\vepsilon}[h(\tilde\vt + \vepsilon) = 1] = p 
    \] yield the same solution.

\end{lemma}

\begin{proof}
    The term $\upperval[\lowerval[p, \gB], \gB^{-1}]$ is expressed as the following optimization problem:
    \begin{align}
    \label{eq:combined-lbub}
        p^*_\mathrm{high} = \max_{h \in \gH} \Pr_{\vepsilon}[h(\tilde\vt + \vepsilon) = 1]& \quad \text{s.t.} \quad \Pr_{\vepsilon}[h(\vt + \vepsilon) = 1] = p^*_\mathrm{low} \\
        p^*_\mathrm{low} = \min_{h' \in \gH} \Pr_{\vepsilon}[h'(\tilde\vt + \vepsilon) = 1] 
        & \quad \text{s.t.} \quad \Pr_{\vepsilon}[h'(\vt + \vepsilon) = 1] = \Pr_{\vepsilon}[f(\vt + \vepsilon)] = p \nonumber
    \end{align}
    We swap $\tilde\vt$, and $\vt$ since we can use the canonical points interchangeably.
    We have
    \[
     p^*_\mathrm{high} = \max_{h \in \gH} \Pr_{\vepsilon}[h(\vt + \vepsilon) = 1] \quad \text{s.t.} \quad \Pr_{\vepsilon}[h(\tilde\vt + \vepsilon) = 1] = p^*_\mathrm{low}
    \]
    $h^*_\mathrm{low}$ as the solution to the inner problem in \autoref{eq:combined-lbub} (defining $p^*_\mathrm{low}$) is a feasible solution to the outer optimization (the first line); therefore
    \[
        p^*_\mathrm{high} = \max_{h \in \gH} \Pr_{\vepsilon}[h(\vt + \vepsilon) = 1] \ge \Pr_{\vepsilon}[h^*_\mathrm{low}(\vt + \vepsilon) = 1] = \Pr_{\vepsilon}[f(\vt + \vepsilon)] = p
    \]
    Both functions $\upperval$, and $\lowerval$ (and therefore both minimization and maximization) are non-decreasing to the value in their constraint.
    % Since $f$ is a feasible solution for both the inner and outer problem, it implies that $\upperval[\lowerval[p, \gB], \gB^{-1}] \ge p$ as $p$ is one of the feasible solutions. 
    % Consider the specific solution resulting in probability $p$. Let $p_l:= \min_{h' \in \gH} \Pr_{\vepsilon}[h'(\tilde\vx + \vepsilon) = 1]$ on that specific configuration. 
    Assuming $p^*_\mathrm{high} > p$ ($p^*_\mathrm{high} \neq p$), we have $p = \upperval[p'_\mathrm{low}, \gB^{-1}]$ that $p'_\mathrm{low} < p^*_\mathrm{low}$ (due to non-decreasing nature of $\upperval$). We have
    \[
    p = \max_{h \in \gH} \Pr_{\vepsilon}[h(\vt + \vepsilon) = 1] \quad \text{s.t.} \quad \Pr_{\vepsilon}[h(\tilde\vt + \vepsilon) = 1] = p'_\mathrm{low}
    \]
    with a maximizer function $h'_\mathrm{high}$; i.e. $p = \Pr_{\vepsilon}[h'_\mathrm{high}(\vt + \vepsilon) = 1]$.
    We rewrite inner problem in \autoref{eq:combined-lbub} \[
        p^*_\mathrm{low} = \min_{h' \in \gH} \Pr_{\vepsilon}[h'(\tilde\vt + \vepsilon) = 1] 
        \quad \text{s.t.} \quad \Pr_{\vepsilon}[h'(\vx + \vepsilon) = 1] = \Pr_{\vepsilon}[f(\vt + \vepsilon)] = p
    \]
    The maximizer function $h'_\mathrm{high}$ satisfies the constraint, and therefore $p^*_\mathrm{low} < p'_\mathrm{low}$ which is a contradiction. Therefore $p^*_\mathrm{high}=p$.
    % Assuming that there exists a $p^* > p$ as a solution for \autoref{eq:combined-lbub}. Due to the non-decreasing property of the maximization problem, in the new configuration, we have $p_l^+:= \min_{h' \in \gH} \Pr_{\vepsilon}[h'(\tilde\vx + \vepsilon) = 1] > p_l$. Applying the same argument again we have $\Pr_{\vepsilon}[h'(\vx + \vepsilon) = 1] \neq p$ which contradicts the base constraint of the optimization. Therefore $p$ is a feasible solution, and any solution higher than $p$ is not feasible. A similar chain of arguments can be applied to show that $\lowerval[\upperval[p, \gB], \gB^{-1}] = p$.
\end{proof}

\subsection{Lower and Upper Bounds for All Shapes and Sizes}
\label{sec: Lower and Upper Bounds for All Shapes and Sizes}

\begin{lemma}
    \label{thrm:symmetry-lbub}
    For a binary classifier $f(\vx)$, let $g(\vx) = \Pr_{\vepsilon}[f(\vx + \vepsilon) = 1]$ and \[
    g(\tilde\vx) \ge \lowerval_g[p, \gB] := \min_{h \in \gH} \Pr_{\vepsilon}[h(\tilde\vx + \vepsilon) = 1] \quad \text{s.t.} \Pr_{\vepsilon}[h(\vx + \vepsilon) = 1] = g(\vx) = p
    \]
    Similarly let \[
    g(\tilde\vx) \le \upperval_g[p, \gB] := \max_{h \in \gH} \Pr_{\vepsilon}[h(\tilde\vx + \vepsilon) = 1] \quad \text{s.t.} \Pr_{\vepsilon}[h(\vx + \vepsilon) = 1] = g(\vx) = p
    \]
    Both be obtainable at the same canonical points. We have $\upperval_g[p, \gB] = 1 - \lowerval_g[1 - p, \gB]$.
\end{lemma}

\begin{proof}
    For simpler notation let $\overline{g}(\vx) = \upperval_g[g(\vx), \gB]$, $\underline{g}(\vx) = \lowerval_g[g(\vx), \gB]$, then we have
    \begin{align*}
        1 - \overline{g}(\vx) = 1 - \max_{h \in \gH} \Pr_{\vepsilon}[h(\tilde\vx + \vepsilon) = 1] \\
        \text{Let } h'(\vx) = 1 - h(\vx)\text{ then} \\
        = 1 - \max_{h' \in \gH} \Pr_{\vepsilon}[1 - h'(\tilde\vx + \vepsilon) = 1] = \min_{h' \in \gH} \Pr_{\vepsilon}[h'(\tilde\vx + \vepsilon) = 1]
    \end{align*}
    The constraint also translates similarly \begin{align*}
        \Pr_{\vepsilon}[h(\vx + \vepsilon) = 1] = \Pr_{\vepsilon}[1 - h'(\vx + \vepsilon) = 1] = 1 - \Pr_{\vepsilon}[h'(\vx + \vepsilon) = 1] = 1 - p
    \end{align*} And the new problem is by definition same as $1 - \lowerval[1 - p, \gB]$.
\end{proof}

\parbold{Certified Upper and Lower bounds for all Shapes and Sizes} In \autoref{thrm:allshapes} we rephrased the Theorem 4.1 from \citet{yang2020randomized} to return the upper bound probability instead of the robust radius. Here we prove \autoref{thrm:allshapes}, using the original proof from \citet{yang2020randomized}.

Let $g(\vx) := \E_{\vepsilon}[f(\vx + \vepsilon)]$ for any binary decision function $f$ and any $\vepsilon \sim \xi$ where $\xi(\vx)\propto \exp(-\psi(\vx))$. If $g(\cdot)$ is continuous in $\gX$, for any point $\tilde\vx = \vx + \vdelta$ one can compute the $g(\vx + \tilde\vdelta)$ through line integral as \[
g(\tilde\vx) = g(\vx+\vdelta) = g(\vx) + \int_0^r \frac{\mathrm{d}}{\mathrm{dt}}[g(\vx + t\cdot\vdelta')]\mathrm{d}t
\]
where $\vdelta'=\frac{\vdelta}{\|\vdelta\|}$ is the unit vector in the same direction as $\vdelta$ and $r:=\|\vdelta\|$. In other words, we add all the infinitisimal changes on the path from $\vx$ to $\tilde\vx$ to compute the value of $g(\tilde\vx)$ given $g(\vx)$.

With the decision boundary $\gU = \{\vx: f(\vx) = 1\}$, and $\gU - \vz$ as the decision boundary translated by $-\vz$, consider the following function \[
\Omega(p) := \sup_{\vdelta: \|\vdelta\| = 1} \quad \sup_{\gU \in \sR^d: \xi(\gU) = p} \quad \lim_{r \to 0^+} \frac{\xi(\gU - r\vdelta) - p}{r}
\]

Proposition F.8 from \citet{yang2020randomized} show that anywhere in $\gX$, we have $\frac{\mathrm{d}}{\mathrm{dt}}[g(\vz)] \le \Omega(g(\vz))$. This means that one can upper bound the growth of $g(\vx)$ while shifted by $\vdelta$ by integrating over $\Omega(g(\vz))$ instead. For easier notation let $h(t) = g(\vx + t\cdot\vdelta')$ which implies $h'(t) = \frac{\mathrm{d}}{\mathrm{d}t} h(t) \le \Omega(h(t))$.
Notably, the function $\Omega(p)$ is always non-negative. See the equivalent Definition H.12 from \citet{yang2020randomized} where the function is denoted as $\Phi(p)$ and it is written as an expectation of a maximum over a value that is always non-negative. For positive (non zero) $\Omega(h(t))$, including but not limited to $h(t) \le 1/2$ (see the definition of the function in Appendix F from \citet{yang2020randomized}) It follows: \begin{align*}
    \frac{h'(t)}{\Omega(h(t))} \le 1 \Rightarrow \int_{0}^{r} \frac{h'(t)}{\Omega(h(t))}\mathrm{d}t \le \int_0^r 1\mathrm{d}t =r
\end{align*} We set $u = h(t) \Rightarrow \mathrm{d}u = h'(t)\mathrm{d}t$ which implies \begin{align*}
    \Pi(u) = \int_{0}^{r} \frac{h'(t)}{\Omega(h(t))} \mathrm{d}t= \int_{u=h(0)}^{u=h(r)} \frac{1}{\Omega(u)} \mathrm{d}u \le r
\end{align*} 

Here $h(0) = g(\vx + 0 \cdot \vdelta') = g(\vx) = \beta$, and $h(0) = g(\vx + r \cdot \vdelta') = g(\tilde\vx) = \bar\beta$.
With the reference function $F(\gamma) = \int_{\gamma}^{1/2}\frac{1}{\Omega(p)} \mathrm{d}p$ we have 
\[
r \ge \int_{\beta}^{\bar\beta}\frac{1}{\Omega(p)} \mathrm{d}p = \int_{\beta}^{1/2}\frac{1}{\Omega(p)} \mathrm{d}p - \int_{\bar\beta}^{1/2}\frac{1}{\Omega(p)} \mathrm{d}p = F(\beta) - F(\bar\beta) \Rightarrow F(\beta) - F(\bar\beta) \le r
\]  which is $F(\bar\beta) \ge F(\beta) - r$.

The authors already computed $\Omega(p)$ (in their paper it is called as $\Phi$) for following several distributions, including:
\begin{itemize}
    \item Isotropic Gaussian smoothing against $\ell_2$ ball ($\sigma=1$): $\Omega(u) = \Phi'(\Phi^{-1}(1 - u))$ which implies: \begin{align*}    
    \Pi_\mathrm{Gaussian}(u) = \int_{\beta}^{\overline{\beta}} \frac{1}{\Omega(u)}\mathrm{d}u = \int_{\beta}^{\overline{\beta}}\frac{1}{\Phi'(\Phi^{-1}(1-u))}\mathrm{d}u  
    \end{align*} With $c = \Phi^{-1}(1 - u)$ we have $\differ p = - \Phi'(c) \differ c$, and $\Phi'(c) = \Phi'(\Phi^{-1}(1 - u))$. Therefore \begin{align*}
        \int_{\beta}^{\overline{\beta}}\frac{1}{\Phi'(\Phi^{-1}(1-u))}\mathrm{d}u =  \int_{\Phi^{-1}(1 - \beta)}^{\Phi^{-1}(1 - \overline{\beta})} -\differ u = \Phi^{-1}(1 - \beta) - \Phi^{-1} (1 - \overline{\beta}) \le r\\
        \Rightarrow \Phi^{-1}(1 - \overline{\beta}) \ge \Phi^{-1} (1 - \beta) - r \Rightarrow 1 - \Phi^{-1}(\overline{\beta}) \ge 1 - \Phi^{-1}(\beta) - r \\
        \Rightarrow \Phi^{-1} (\overline{\beta}) \le \Phi^{-1}(\beta) + r \rightarrow \overline{\beta} \le \Phi(\Phi^{-1}(\beta) + r)
    \end{align*} Which completely aligns with the aforementioned closed form $\ell_2$ certificate.
    \item Laplace smoothing against $\ell_1$ ball ($\sigma = \sqrt{2}\lambda$): $\Omega(u) = \frac{u}{\lambda}$ which implies:
    \begin{align*}
        \Pi_\mathrm{Laplace}(u) = \int_{\beta}^{\overline{\beta}} \frac{1}{\Omega(u)}\mathrm{d}u = \int_{\beta}^{\overline{\beta}} \lambda\frac{1}{u}\mathrm{d}u = \lambda\log\frac{\overline{\beta}}{\beta} \le r \\
        \Rightarrow \frac{\overline{\beta}}{\beta} \le 2^{r/\lambda} \Rightarrow \overline{\beta} \le 2^{r/\lambda} \cdot \beta
    \end{align*}

\end{itemize}
Note that in both cases, the function $\Omega(p)$ is positive in $(0, 1)$. 

% An easy way to obtain an upper bound is to use the robust radius (the closed form solution for $\int_{\beta}^{1/2} \frac{1}{\Omega(p)} \differ p$) as a proxy. We have $\int_{\beta}^{\overline{\beta}} \frac{1}{\Omega(p)} \differ p = \int_{\beta}^{1/2} \frac{1}{\Omega(p)} \differ p - \int_{\overline{\beta}}^{1/2} \frac{1}{\Omega(p)} \differ p$. This function is monotonically increasing in $[0, 1/2]$, which means that we can use binary search to find the upper bound $\overline{\beta}$ as:
% \begin{align}
%     \overline{\beta} = \sup \left\{ \hat\beta \in [\beta, 0.5]: \int_{\hat{\beta}}^{1/2} \frac{1}{\Omega(p)} \differ p \ge \int_{{\beta}}^{1/2} \frac{1}{\Omega(p)} \differ p - r\right\}
% \end{align}

\section{Supplementary Experiments}
\label{sec: Supplementary Experiments}

\parbold{Compute resources} We ran our experiment using Nividia A-100 and H-100 Tensor Core GPUs. For each experiment only one GPU was used. We use the A-100 GPU for the CIFAR-10 dataset under ResNet setup, and the conformal risk control experiment. The rest of the results use H-100 as the compute resource.

\parbold{Experimental setup} For the CIFAR-10 datasets we evaluate the results over 2048 test samples for ResNet model and 10000 images for the ViT models. For the ImageNet since the number of classes are 1000, we report our results over 5000 images for ViT models and 50000 images on ResNet models. Ultimately the number of samples does not influence the empirical results. The number of Monte Carlo samples are initially set to 500 for CIFAR and 300 for ImageNet. For each experiment, and for the reported sample rate we cut the precomputed samples, from the reported number. 

Our results are reports over 100 runs (except the conformal risk control which is over one run. In each run we sample the 10\% of the points as the calibration set. For conformal risk control we report the result on 300 images where 100 random images from it is taken for the calibration. Ultimately the size of the calibration set does not effect the final performance. As the calibration set gets larger the distribution of the coverage probability concentrates around $1 - \alpha$. 

The time to compute the logits for the CIFAR-10 dataset is 1:30:56 (ViT with 10000 datapoints and 500 samples), and for the ImageNet dataset it is 13:52:11 (ViT with 5000 datapoints and 300 MC samples). For the ImageNet, and the ResNet model this number is 2:52:11 (for 50000 datapoints and 1000 samples).

\begin{figure}
    \centering
    \input{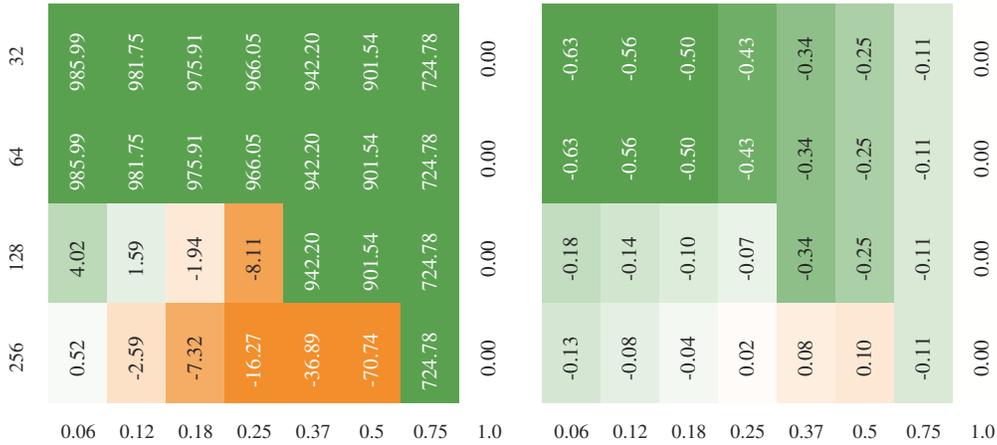}
    \caption{[Left] Comparison of the average set size $|\gC_{r, \mathrm{BinCP}}| - |\gC_{r, \mathrm{\method}}|$ and [Right] the proportion of the sets with size $\le 10$ expressed in BinCP - \method. In both plots green shows that \method is performing better. To convert the relative difference to absolute number \autoref{tab:imagenet-resnet-ref} can be used as the reference.}
    \label{fig:imagenet-size-heatmap}
\end{figure}

\parbold{Set size experiment}
Tables \autoref{tab:cifar-full-25} and \autoref{tab:cifar-full-5} report the empirical coverage and average prediction set size of \method for different radii $r$ on the CIFAR-10 dataset using a ResNet model under two noise levels, $\sigma = 0.25$ and $\sigma = 0.5$, respectively. We also report the result of the ImageNet dataset (for the ResNet model) in \autoref{tab:imagenet-resnet-ref}. Specifically for this dataset, because of the large number of classes we also reported the proportion of the sets below specific sizes (1, 3, 5, and 10). As expected, increasing the radius $r$ results in a more conservative setup and hence higher coverage on the clean points. 
For CIFAR-10 dataset \autoref{fig:cifar-all-resnet}, and for the ImageNet dataset \autoref{fig:imagenet-size-heatmap} visualize the comparative performance between BinCP and RCP1 across various radii and sampling budgets. These results are on the ResNet model. The heatmaps show the difference in set sizes $|\mathcal{C}_{r,\text{BinCP}}| - |\mathcal{C}_{r,\text{RCP1}}|$, where positive (green) values indicate that RCP1 provides smaller or more efficient sets. RCP1 generally outperforms BinCP across low sample rates, especially for smaller radii and moderate sampling budgets. The reference tables (Tables \autoref{tab:cifar-full-25} and \autoref{tab:cifar-full-5}, \autoref{tab:imagenet-resnet-ref}) can be used to interpret these differences in absolute terms.

\begin{figure}
    \centering
    \input{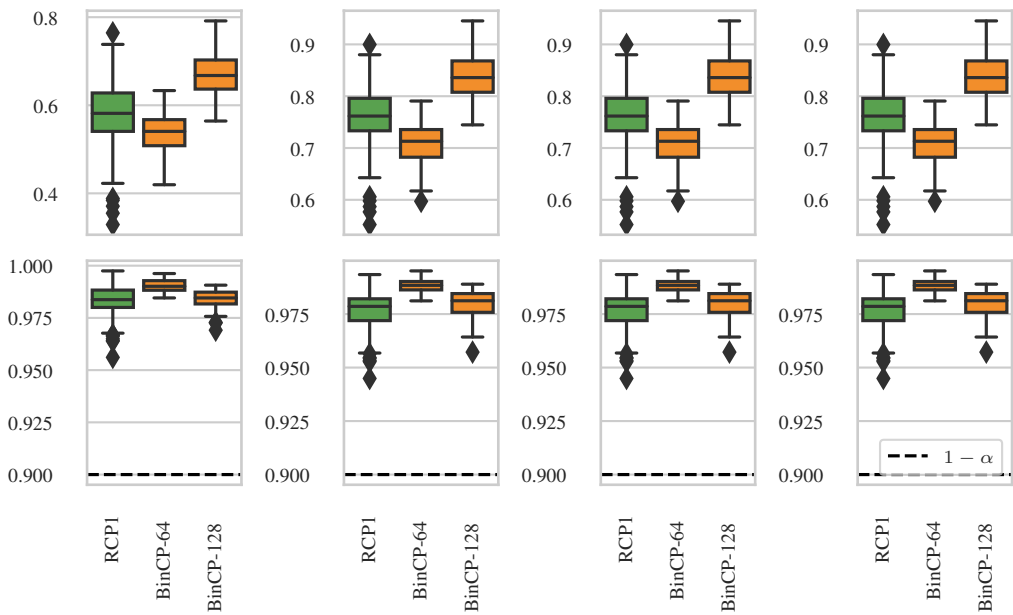}
    \caption{[Up] The proportion and [Bottom] the coverage of the prediction sets with size [From left to right] $|\gC| \le 1$, $|\gC| \le 3$, $|\gC| \le 5$, $|\gC| \le 10$.}
    \label{fig:small-sets-cov}
\end{figure}

\parbold{Proportion of small sets} As also discussed in \autoref{sec: Experiments} although the proportion of sets with size less than a threshold shows how applicable a CP algorithm is, it can be misleading -- a CP framework can return many false prediction sets with very small set size. Therefore alongside the proportion of these sets we should also report their coverage. We show these results in \autoref{fig:small-sets-cov}. Our observation is that all setups result in sets with coverage higher than the determined level. Note that in terms of proportion \method stands somewhere between BinCP with 64 and 128 samples which aligns with our aforementioned intuition.

\begin{figure}
    \begin{minipage}[t]{0.48\textwidth}
    % \begin{table}[t]
    \centering
    \captionsetup{type=table} % needed for caption outside float
    \captionof{table}{Empirical coverage and average set size for different radii ($r$), for CIFAR-10 dataset with ResNet model and $\sigma=0.25$.}
    % \caption{}
    \vspace{0.5em}
    
    \begin{tabular}{@{}ccc@{}}
    \toprule
    \rowcolor{gray!10}
    \textbf{$r$} & \textbf{Empirical Coverage} & \textbf{Avg Set Size} \\
    \midrule
    0.06 & 0.936 $\pm$ 0.018 & 2.156 $\pm$ 0.241 \\
    0.12 & 0.961 $\pm$ 0.014 & 2.646 $\pm$ 0.306 \\
    0.18 & 0.981 $\pm$ 0.010 & 3.315 $\pm$ 0.478 \\
    0.25 & 0.990 $\pm$ 0.008 & 4.178 $\pm$ 0.798 \\
    0.37 & 1.000 $\pm$ 0.000 & 10.000 $\pm$ 0.000 \\
    0.50 & 1.000 $\pm$ 0.000 & 10.000 $\pm$ 0.000 \\
    0.75 & 1.000 $\pm$ 0.000 & 10.000 $\pm$ 0.000 \\
    \bottomrule
    \end{tabular}
    
    \vspace{-0.5em}
    \label{tab:cifar-full-25}
    % \end{table}
    \end{minipage}
    \hfill
    \begin{minipage}[t]{0.48\textwidth}
    \centering
    \captionsetup{type=table} % needed for caption outside float
    \captionof{table}{Empirical coverage and average set size for different radii ($r$), for CIFAR-10 dataset with ResNet model and $\sigma=0.5$.}
    \vspace{0.5em}
    
    \begin{tabular}{@{}ccc@{}}
    \toprule
    \rowcolor{gray!10}
    \textbf{$r$} & \textbf{Empirical Coverage} & \textbf{Avg Set Size} \\
    \midrule
    0.06 & 0.921 ± 0.020 & 2.684 ± 0.244 \\
    0.12 & 0.937 ± 0.018 & 2.937 ± 0.285 \\
    0.18 & 0.951 ± 0.016 & 3.236 ± 0.356 \\
    0.25 & 0.966 ± 0.014 & 3.741 ± 0.530 \\
    0.37 & 0.980 ± 0.010 & 4.500 ± 0.560 \\
    0.50 & 0.990 ± 0.007 & 5.300 ± 0.712 \\
    0.75 & 1.000 ± 0.000 & 10.000 ± 0.000 \\
    \bottomrule
    \end{tabular}
    
    \vspace{-0.5em}
    \label{tab:cifar-full-5}
    \end{minipage}
\end{figure}

% \begin{table}[t]

% \end{table}

\begin{figure}
    \centering
    \input{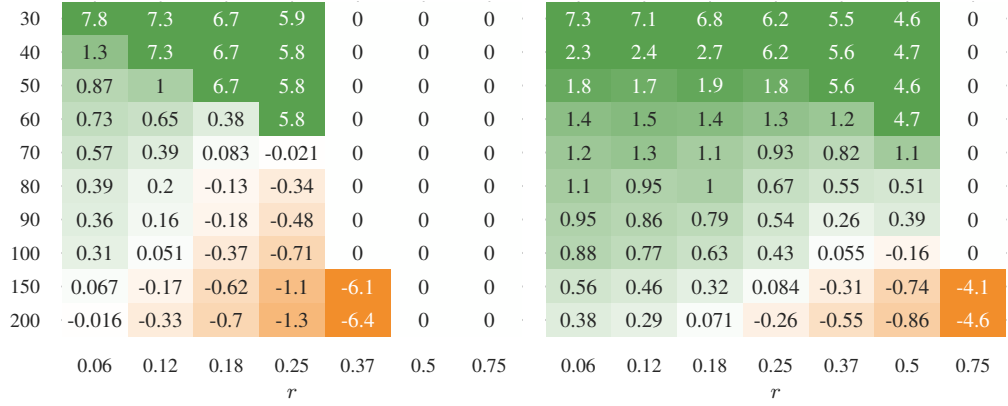}
    
    \caption{Comparison of BinCP and \method in terms of $|\gC_{r, \mathrm{BinCP}}| - |\gC_{r, \mathrm{\method}}|$ (higher (green) shows better performance for \method) across various radii and sample rates. Results are on the ResNet model and for the CIFAR-10 dataset. [Left] $\sigma=0.25$, and [Right] $\sigma=0.5$. Note that the numbers are in terms of difference and to compute the absolute number \autoref{tab:cifar-full-25}, and \autoref{tab:cifar-full-5} can be used as reference.}
    \label{fig:cifar-all-resnet}
\end{figure}

\begin{table}[t]
\centering
\caption{Statistics from \method across various radii. The results are for ImageNet dataset and the ResNet model.}
\vspace{0.5em}
\begin{adjustbox}{width=0.99\linewidth}
\begin{tabular}{@{}ccccccc@{}}
\toprule
\rowcolor{gray!10}
\textbf{$r$} & \textbf{Avg Set Size} & \textbf{Emp. Coverage} & \textbf{$\gC\le 1$} & \textbf{$\gC\le 3$} & \textbf{$\gC\le 5$} & \textbf{$\gC\le 10$} \\
\midrule
0.06 & 14.013 $\pm$ 1.787 & 0.921 $\pm$ 0.008 & 0.139 $\pm$ 0.010 & 0.311 $\pm$ 0.018 & 0.430 $\pm$ 0.025 & 0.626 $\pm$ 0.035 \\
0.12 & 18.246 $\pm$ 2.606 & 0.936 $\pm$ 0.009 & 0.120 $\pm$ 0.010 & 0.274 $\pm$ 0.019 & 0.383 $\pm$ 0.024 & 0.560 $\pm$ 0.032 \\
0.18 & 24.095 $\pm$ 3.645 & 0.951 $\pm$ 0.008 & 0.101 $\pm$ 0.010 & 0.239 $\pm$ 0.018 & 0.336 $\pm$ 0.023 & 0.501 $\pm$ 0.031 \\
0.25 & 33.953 $\pm$ 5.744 & 0.964 $\pm$ 0.007 & 0.082 $\pm$ 0.008 & 0.201 $\pm$ 0.017 & 0.288 $\pm$ 0.022 & 0.432 $\pm$ 0.033 \\
0.37 & 57.802 $\pm$ 10.753 & 0.979 $\pm$ 0.005 & 0.058 $\pm$ 0.008 & 0.151 $\pm$ 0.017 & 0.219 $\pm$ 0.022 & 0.337 $\pm$ 0.031 \\
0.50 & 98.464 $\pm$ 19.136 & 0.989 $\pm$ 0.003 & 0.036 $\pm$ 0.006 & 0.104 $\pm$ 0.016 & 0.160 $\pm$ 0.021 & 0.252 $\pm$ 0.029 \\
0.75 & 275.222 $\pm$ 88.968 & 0.998 $\pm$ 0.002 & 0.012 $\pm$ 0.006 & 0.036 $\pm$ 0.016 & 0.063 $\pm$ 0.025 & 0.115 $\pm$ 0.039 \\
1.00 & 1000.000 $\pm$ 0.000 & 1.000 $\pm$ 0.000 & 0.000 $\pm$ 0.000 & 0.000 $\pm$ 0.000 & 0.000 $\pm$ 0.000 & 0.000 $\pm$ 0.000 \\
\bottomrule
\end{tabular}
\end{adjustbox}
\vspace{-0.5em}
\label{tab:imagenet-resnet-ref}
\end{table}

\paragraph{Regression Experiment}
For robust conformal regression with \method we use the Udacity and originates from Nvidia’s DAVE-2 system\citep{bojarski2016end}. The input of this task is an scene, and the task is to estimate the steering angle of the car. The output range is from -1 (completely steering right) to 1 (left). For this task we finetune a ResNet18 model \citep{he2016deep} on images augmented with isotopic Gaussian noise with $\sigma=0.5$. We run finetuning for 200 epochs. We use the same $\sigma$ for augmenting the input in \method. We set $1 - \alpha=0.9$, and evalute on $r \in \{0.12, 0.25, 0.5\}$. To the best of our knowledge our result is the first robust conformal prediction with randomized smoothing for regression task. \autoref{tab:regression} compares the interval length and empirical coverage across various radii. 
\begin{table}
\centering
\caption{\method for conformal regression. We report the empirical coverage and the interval length across varius radii.}
\label{tab:regression}
\begin{tabular}{@{}ccc@{}}
\toprule
\rowcolor{gray!10}
\textbf{$r$} & \textbf{Empirical Coverage} & \textbf{Interval Width} \\
\midrule
0.00 & 0.900 $\pm$ 0.005 & 0.371 $\pm$ 0.012 \\
0.12 & 0.920 $\pm$ 0.005 & 0.426 $\pm$ 0.014 \\
0.25 & 0.938 $\pm$ 0.004 & 0.494 $\pm$ 0.018 \\
0.50 & 0.963 $\pm$ 0.003 & 0.667 $\pm$ 0.026 \\
\bottomrule
\end{tabular}

\end{table}
% \begin{figure}
%     \centering
%     \input{figures/cifar_setsize_heatmap_original_models.pgf}
%     \caption{Comparison of set size between BinCP, and \method on CIFAR-10 dataset with original baseline used by \citet{anonymous2024robust}. The results are over various sample rates and radii for Gaussian smoothing, $\ell_2$ certificate respectively with [Left] $\sigma=0.12$, [Middle] $\sigma=0.25$, and [Right] $\sigma=0.5$. The heatmap shows the set size of $($ BinCP $-$ \method$)$}
%     \label{fig:enter-label}
% \end{figure}

% \begin{figure}
%     \centering
%     \input{figures/imagenet_setsize_heatmap.pgf}
%     \caption{[Top-left] Comparison of average set size, and the proportion of set sizes $\le 1$ [Top-middle], $\le 3$ [Top-right], $\le 5$ [Bottom-left], $\le 10$ [Bottom-middle], and $\le 20$ [Bottom-right] for the ImageNet dataset with Gaussian smoothing on the baseline setup with $\sigma=0.5$. The results are shown in terms of ( BinCP - \method) for various sample-rates (only applies to BinCP) and radii.}
%     \label{fig:enter-label}
% \end{figure}

\newpage

\section*{NeurIPS Paper Checklist}

\begin{enumerate}

\item {\bf Claims}
    \item[] Question: Do the main claims made in the abstract and introduction accurately reflect the paper's contributions and scope?
    \item[] Answer: \answerYes{} % Replace by \answerYes{}, \answerNo{}, or \answerNA{}.
    \item[] Justification: Other claims that are already proven in prior works we prove every new claim.

\item {\bf Limitations}
    \item[] Question: Does the paper discuss the limitations of the work performed by the authors?
    \item[] Answer: \answerYes{} % Replace by \answerYes{}, \answerNo{}, or \answerNA{}.
    \item[] Justification: We discussed the limitation of our work at the end of \autoref{sec: Experiments}.

\item {\bf Theory assumptions and proofs}
    \item[] Question: For each theoretical result, does the paper provide the full set of assumptions and a complete (and correct) proof?
    \item[] Answer: \answerYes{} % Replace by \answerYes{}, \answerNo{}, or \answerNA{}.
    \item[] Justification: We dedicated \autoref{sec:proofs}, and \autoref{sec: Lower and Upper Bounds for All Shapes and Sizes} for our proofs and referenced them in the manuscript.

    \item {\bf Experimental result reproducibility}
    \item[] Question: Does the paper fully disclose all the information needed to reproduce the main experimental results of the paper to the extent that it affects the main claims and/or conclusions of the paper (regardless of whether the code and data are provided or not)?
    \item[] Answer: \answerYes{} % Replace by \answerYes{}, \answerNo{}, or \answerNA{}.
    \item[] Justification: Both in the last paragraph of the introduction (\autoref{sec: Introduction}) and \autoref{sec: Robust CP with a Single Sample} the procedure is explained. Datasets, models, and evaluation criteria are also discussed in \autoref{sec: Experiments}

\item {\bf Open access to data and code}
    \item[] Question: Does the paper provide open access to the data and code, with sufficient instructions to faithfully reproduce the main experimental results, as described in supplemental material?
    \item[] Answer: \answerYes{} % Replace by \answerYes{}, \answerNo{}, or \answerNA{}.
    \item[] Justification: Our code is uploaded as a zip file in the supplementary materials. After acceptance we also share the code in GitHub. All datasets are well-known in the literature, and all models are pretrained and open source. We also cited the original work introducing that dataset or model.

\item {\bf Experimental setting/details}
    \item[] Question: Does the paper specify all the training and test details (e.g., data splits, hyperparameters, how they were chosen, type of optimizer, etc.) necessary to understand the results?
    \item[] Answer: \answerYes{} % Replace by \answerYes{}, \answerNo{}, or \answerNA{}.
    \item[] Justification: We discussed the parameters in \autoref{sec:fig-manual}, and in the caption of the figures.

\item {\bf Experiment statistical significance}
    \item[] Question: Does the paper report error bars suitably and correctly defined or other appropriate information about the statistical significance of the experiments?
    \item[] Answer: \answerYes{} % Replace by \answerYes{}, \answerNo{}, or \answerNA{}.
    \item[] Justification: We show the confidence intervals as error bars in the plots. Later in \autoref{tab:tps_aps_full} we represent the results in form of mean $\pm$ std.

\item {\bf Experiments compute resources}
    \item[] Question: For each experiment, does the paper provide sufficient information on the computer resources (type of compute workers, memory, time of execution) needed to reproduce the experiments?
    \item[] Answer: \answerYes{} % Replace by \answerYes{}, \answerNo{}, or \answerNA{}.
    \item[] Justification: We discussed the compute resources in \autoref{sec: Supplementary Experiments}. For the wall-clock time report in \autoref{tab:smoothing_runtime} we explicitly reported the resource in the caption.
    
\item {\bf Code of ethics}
    \item[] Question: Does the research conducted in the paper conform, in every respect, with the NeurIPS Code of Ethics \url{https://neurips.cc/public/EthicsGuidelines}?
    \item[] Answer: \answerYes{} % Replace by \answerYes{}, \answerNo{}, or \answerNA{}.
    \item[] Justification: We did not involve any human participation in the experiments. All datasets and models are open source and known to the literature.

\item {\bf Broader impacts}
    \item[] Question: Does the paper discuss both potential positive societal impacts and negative societal impacts of the work performed?
    \item[] Answer: \answerNA{} % Replace by \answerYes{}, \answerNo{}, or \answerNA{}.
    \item[] Justification: The work is a fundamental reseach and does not have potential societal impacts.
    
\item {\bf Safeguards}
    \item[] Question: Does the paper describe safeguards that have been put in place for responsible release of data or models that have a high risk for misuse (e.g., pretrained language models, image generators, or scraped datasets)?
    \item[] Answer: \answerNA{} % Replace by \answerYes{}, \answerNo{}, or \answerNA{}.
    \item[] Justification: The work is a fundamental AI research and poses no such risks. 

\item {\bf Licenses for existing assets}
    \item[] Question: Are the creators or original owners of assets (e.g., code, data, models), used in the paper, properly credited and are the license and terms of use explicitly mentioned and properly respected?
    \item[] Answer: \answerYes{} % Replace by \answerYes{}, \answerNo{}, or \answerNA{}.
    \item[] Justification: We have cited all the datasets and models used in the work.

\item {\bf New assets}
    \item[] Question: Are new assets introduced in the paper well documented and is the documentation provided alongside the assets?
    \item[] Answer: \answerNA{} % Replace by \answerYes{}, \answerNo{}, or \answerNA{}.
    \item[] Justification: We do not release any new assets.

\item {\bf Crowdsourcing and research with human subjects}
    \item[] Question: For crowdsourcing experiments and research with human subjects, does the paper include the full text of instructions given to participants and screenshots, if applicable, as well as details about compensation (if any)? 
    \item[] Answer: \answerNA{} % Replace by \answerYes{}, \answerNo{}, or \answerNA{}.
    \item[] Justification: The paper does not involve crowdsourcing nor research with human subjects.

\item {\bf Institutional review board (IRB) approvals or equivalent for research with human subjects}
    \item[] Question: Does the paper describe potential risks incurred by study participants, whether such risks were disclosed to the subjects, and whether Institutional Review Board (IRB) approvals (or an equivalent approval/review based on the requirements of your country or institution) were obtained?
    \item[] Answer: \answerNA{} % Replace by \answerYes{}, \answerNo{}, or \answerNA{}.
    \item[] Justification: the paper does not involve crowdsourcing nor research with human subjects.

\item {\bf Declaration of LLM usage}
    \item[] Question: Does the paper describe the usage of LLMs if it is an important, original, or non-standard component of the core methods in this research? Note that if the LLM is used only for writing, editing, or formatting purposes and does not impact the core methodology, scientific rigorousness, or originality of the research, declaration is not required.
    %this research? 
    \item[] Answer: \answerNA{} % Replace by \answerYes{}, \answerNo{}, or \answerNA{}.
    \item[] Justification: We only used LLMs in some parts to improve the quality of the text.

\end{enumerate}

\end{document}